\documentclass[10pt,leqno]{amsart}

\usepackage{graphicx}
\usepackage{amsmath,amssymb,amsthm}
\usepackage{capt-of}
\usepackage{xcolor}
\usepackage{indentfirst,csquotes}
\usepackage[authoryear,round]{natbib}
\usepackage[hidelinks]{hyperref}

\AtBeginDocument{\setlength{\baselineskip}{16pt}}

\title[Differentiable time-lapse ERT inversion]
{An automatic-differentiation framework for time-lapse electrical
resistivity tomography inversion of hydrologic dynamics}

\author{Pu Yang$^{1}$, Zhengyang Fang$^{1,2}$, Yuxin Liu$^{1}$,
Xuan Su$^{3}$, Deshan Feng$^{1}$, and Hang Chen$^{2}$}

\address[1]{School of Geosciences and Info-physics, Central South University,
Changsha 410083, China}
\address[2]{School of Earth, Environment, and Sustainability, University of Iowa,
Iowa City, IA 52245, USA}
\address[3]{School of Management Science and Engineering, Hunan University of
Technology and Business, Changsha 410205, China}
\email{hchen117@uiowa.edu}
\thanks{Corresponding author: Hang Chen, hchen117@uiowa.edu}

\date{}

\begin{document}

\begin{abstract}

 {Time-lapse electrical resistivity tomography (TL-ERT) provides spatially distributed information on subsurface hydrologic changes. However, inversion of long monitoring sequences is computationally demanding. Modifying the data misfit, regularization, model parameterization, or petrophysical transformation may also require new gradient derivations and separate implementations. Here, we present AD-TLERT, a unified, GPU-accelerated framework for time-lapse ERT inversion based on automatic differentiation. The framework integrates model parameterization, differentiable petrophysical transformations, forward modeling, data misfit, regularization and auxiliary constraints into a single computational chain. Alternative inversion formulations can therefore reuse the same PDE derivative implementation without re-deriving the complete ERT sensitivity for each case. Comparisons with pyGIMLi showed close agreement in the forward responses, gradients, and recovered resistivity models. Under the tested configuration, AD-TLERT achieved an approximately 51-fold speedup. Synthetic experiments showed that inversion choices affect the amplitude, geometry, and temporal behavior of recovered anomalies. By propagating gradients through the embedded petrophysical relationship, AD-TLERT enabled direct water-content inversion and yielded more accurate estimates than post-inversion conversion for the tested model. A field application further demonstrated how ERT, temperature, and soil-moisture observations can be combined to image snowmelt-driven hillslope wetting. AD-TLERT provides an efficient and flexible framework for time-lapse ERT inversion and hydrologic interpretation.} 

\end{abstract}

\maketitle

\section{Introduction}

 {
Time-lapse electrical resistivity tomography (TL-ERT) is widely used to monitor dynamic subsurface processes, including water infiltration, groundwater flow, solute transport, freeze--thaw cycles, and thermal--hydrological--mechanical processes \citep{Daily1992,Kemna2002,Cassiani2009,Krautblatter2010,Chambers2014,Uhlemann2017,Chen2024Salt,Chen2025Bentonite}. Repeated surveys provide spatially distributed observations of electrical changes that are sensitive to pore-fluid content, salinity, temperature, and lithology \citep{Singha2015,Binley2015}. These observations complement point-based hydrological and environmental measurements \citep{Chen2022Review}. In many applications, however, the main interest is not the absolute resistivity at an individual time, but the spatial and temporal evolution of process-related variables such as water content or saturation \citep{Binley2002,Tso2019}. Time-lapse ERT inversion therefore needs to recover electrical changes and relate them to physically meaningful subsurface states. At larger spatial and temporal scales, process simulations can further guide the interpretation of long-term electrical monitoring data \citep{Robinson2024}.
}

 {
This task becomes computationally demanding as monitoring sequences grow. ERT inversion is nonlinear and ill-posed, and the time-dependent signal may be small relative to the background resistivity structure \citep{LaBrecque1996}. Repeated surveys are also affected by measurement noise, systematic errors associated with changing contact conditions and electrode positions, and environmental variability \citep{Tso2017,Oldenborger2005}. Existing time-lapse inversion methods can be broadly grouped into difference-based and temporally coupled approaches. Difference-based inversion estimates changes relative to a baseline data set or model and can incorporate spatial covariance priors \citep{LaBrecque2001,Hermans2016}. Temporally coupled inversion jointly estimates multiple surveys with constraints between successive models, including simultaneous and full-sequence 4-D formulations \citep{Kim2009,Hayley2011,Loke2014}. Later extensions introduced adaptive temporal weighting, $L_p$-norm penalties, and combined spatial and temporal constraints \citep{Karaoulis2011,Kim2013,Karaoulis2014}. For long records, windowed inversion restricts this coupling to short, overlapping subsets, reducing the size of each inverse problem while preserving local temporal consistency \citep{Wilkinson2022}. Nevertheless, both full-sequence and windowed formulations require repeated forward and sensitivity calculations across many surveys, current sources, and wavenumbers. The resulting computational cost can limit the inversion of dense monitoring data and the systematic evaluation of alternative formulations.
}

 {
The recovered changes also depend strongly on how the inverse problem is formulated. The data-misfit function determines how measurement errors and large residuals influence the model update \citep{Farquharson1998}. Spatial and temporal regularization affect the amplitude, geometry, and smoothness of recovered anomalies \citep{Constable1987,Portniaguine1999,Fiandaca2015}. Optimization methods influence convergence and computational cost, while model parameterization determines whether the inversion is performed in resistivity, conductivity, relative change, or a process-related variable. No inversion setting is universally optimal because the appropriate formulation depends on data quality, survey coverage, expected anomaly geometry, and prior information \citep{Caterina2014,Hermans2016}. Established frameworks such as SimPEG provide modular components for geophysical simulation and inversion \citep{Cockett2015}. Nevertheless, extending a time-lapse ERT implementation to new spatiotemporal objectives, differentiable petrophysical mappings, and GPU sparse solvers can still require problem-specific derivative and implementation work. Changing one component may require a new gradient derivation or a separate implementation, making controlled comparison difficult. A related limitation arises in hydrologic interpretation. Resistivity is commonly inverted first and then converted to water content or other parameters \citep{Tso2019}. Because data fitting and regularization are applied in resistivity space, smoothing and amplitude bias may propagate nonlinearly into the converted estimate. Joint inversion of complementary geophysical data can improve subsurface characterization \citep{Linde2006}, whereas coupled hydrogeophysical inversion can estimate hydrologic states or parameters more directly by embedding process and petrophysical models within the inversion \citep{Hinnell2010,Pleasants2022}. Borehole measurements, sensor observations, and other auxiliary data can likewise be introduced as constraints when a suitable mapping to the inversion variables is available, but each added constraint may require a dedicated coupling term and its derivative. 
}

 {
Automatic differentiation (AD) provides a route to a more flexible implementation \citep{Baydin2018,Margossian2019}. When model parameterization, forward modeling, data misfit, regularization and auxiliary constraints are expressed as differentiable operations, gradients can be assembled consistently through the computational chain. For sparse finite-element problems, AD can be combined with implicit differentiation of the linear-system solution, avoiding differentiation through the internal factorization steps. GPU-based sparse linear algebra can further accelerate the repeated forward and adjoint solves required by time-lapse inversion. Alternative misfit functions, constraints, optimization strategies, and petrophysical parameterizations can then be evaluated within the same forward and gradient implementation \citep{Sambridge2007,Liu2025}. Although AD has recently been applied to ERT inversion \citep{Rincon2026}, its integration with GPU-accelerated 2.5-D time-lapse inversion, systematic evaluation of inversion choices, direct inversion of hydrologic variables and flexible incorporation of auxiliary observations remains limited.
}

 {
In this study, we develop AD-TLERT, a GPU-accelerated differentiable framework for the inversion of hydrologic dynamics in time-lapse ERT. The framework combines a finite-element forward solver, implicit differentiation, GPU sparse linear algebra, temporal coupling, spatial and temporal regularization, auxiliary constraints, and flexible model transformations. We first benchmark its forward responses, gradients, recovered resistivity models, and computational performance against pyGIMLi \citep{Rucker2017}. We then compare representative data-misfit functions, regularization strategies, and optimization algorithms under a common synthetic setting. A differentiable petrophysical relationship based on established conductivity--saturation formulations \citep{Archie1942,Waxman1968} is further embedded in the inversion to compare direct water-content inversion with conventional post-inversion conversion. Finally, the framework is applied to field observations of snowmelt-driven hillslope wetting using temperature and soil-moisture observations as supporting constraints. It provides an efficient and extensible framework for testing inversion assumptions and interpreting hydrologic dynamics.
}

\begin{figure}
\centering
\includegraphics[width=\textwidth]{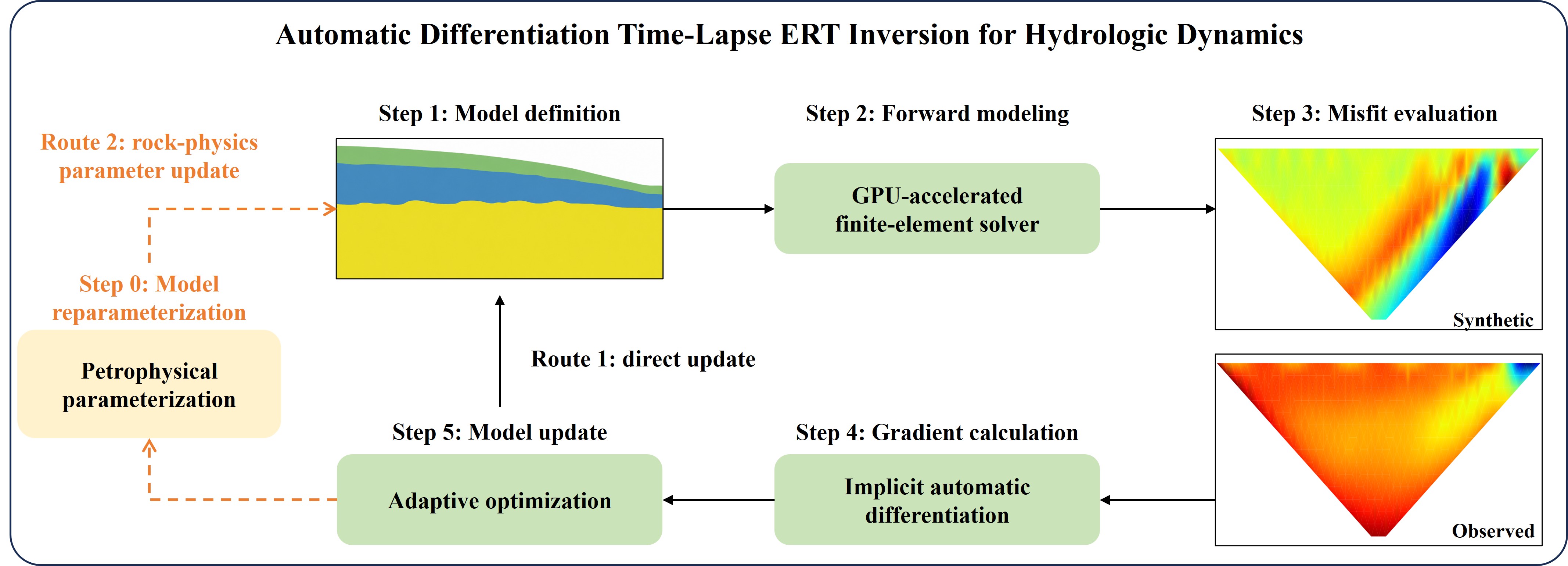}
\caption{Workflow of the automatic-differentiation framework for time-lapse ERT inversion of hydrologic dynamics. Step 1 defines the subsurface model and survey configuration. Step 2 calculates synthetic ERT data using a GPU-accelerated finite-element solver. Step 3 evaluates the discrepancy between the observed and synthetic data using the selected objective function. Step 4 computes the gradients of the objective function with respect to the inversion variables through implicit automatic differentiation. Step 5 updates the model using an adaptive optimization scheme. In Route 1, the resistivity model is updated directly. In the optional Route 2, the resistivity model is expressed through a petrophysical parameterization, and the corresponding rock-physics parameters are updated during inversion.}
\label{fig1}
\end{figure}

\section{Methodology}

This section presents the mathematical formulation and detailed derivations underlying AD-TLERT. As summarized in Fig.~\ref{fig1}, the framework organizes time-lapse inversion into five differentiable stages: model definition, GPU-accelerated finite-element forward modeling, misfit evaluation, gradient calculation by implicit automatic differentiation, and adaptive model updating. Within this modular computational graph, alternative objective functions, optimization algorithms, and spatial and temporal regularization strategies can be selected and combined without re-deriving the complete ERT sensitivity for each inversion formulation. The framework also supports two update routes, which are direct inversion of the resistivity model and an optional petrophysical route in which hydrologic or rock-physics parameters are mapped to resistivity and updated through the same gradient chain. The following subsections formulate the time-lapse inverse problem, derive the implicit gradient of the 2.5-D ERT forward operator, and describe the differentiable petrophysical parameterization.

\subsection{Differentiable time-lapse ERT inversion}

Electrical resistivity tomography (ERT) forward modeling maps a subsurface conductivity distribution to the corresponding survey response \citep{Dey1979}. For conductivity $\sigma$, the electric potential $u$ satisfies
\begin{equation}
-\nabla\cdot\left(\sigma\nabla u\right)=q,
\label{eq:ert_pde}
\end{equation}
where $q$ denotes the impressed current source. After discretization and projection to the receiver locations, the predicted data are written as
\begin{equation}
\mathbf{d}_{\mathrm{pred}}=\mathcal{F}(\mathbf{m}),
\label{eq:forward}
\end{equation}
where $\mathbf{m}=\{\mathbf{m}_t\}_{t=1}^{N_t}$ contains the model parameters at all survey times. The time-lapse inverse problem is formulated as
\begin{equation}
\begin{aligned}
\min_{\mathbf{m}}\ \Phi(\mathbf{m})
={}&
\sum_{t=1}^{N_t}
\Phi_d^{(t)}
\left(
\mathcal{F}(\mathbf{m}_t),
\mathbf{d}_{\mathrm{obs}}^{(t)}
\right)
+
\alpha_s
\sum_{t=1}^{N_t}
\Phi_s(\mathbf{m}_t)
\\
&+
\alpha_t
\sum_{t=2}^{N_t}
\Phi_t(\mathbf{m}_t,\mathbf{m}_{t-1})
+
\alpha_c\Phi_c(\mathbf{m}),
\end{aligned}
\label{eq:objective}
\end{equation}
where $\Phi_d$, $\Phi_s$, and $\Phi_t$ denote the data-misfit, spatial-regularization, and temporal-regularization terms, respectively. The optional term $\Phi_c$ incorporates additional constraints, such
as borehole or sensor observations, petrophysical priors, or coupling
terms used in joint inversion.

All model transformations and the ERT forward operator are included in a differentiable computational graph. The gradient
\begin{equation}
\mathbf{g}=\nabla_{\mathbf{m}}\Phi(\mathbf{m})
\label{eq:gradient}
\end{equation}
is evaluated by automatic differentiation, while the finite-element linear solve is differentiated implicitly. This separation allows the same PDE derivative to be used with different data-misfit functions, regularization terms, and model parameterizations.

\subsection{Implicit differentiation of the 2.5-D ERT forward operator}

We assume that conductivity varies in the survey plane and is invariant in the strike direction, such that $\sigma=\sigma(x,z)$. Applying a cosine transform in the strike direction reduces the three-dimensional governing equation to a set of independent two-dimensional problems,
\begin{equation}
-\nabla_{xz}\cdot
\left(
\sigma\nabla_{xz}\widehat{u}_{k,s}
\right)
+
k^2\sigma\widehat{u}_{k,s}
=
\widehat{q}_{k,s},
\label{eq:wavenumber_pde}
\end{equation}
where $k$ is the wavenumber and $s$ indexes a unit-current pole source. Using first-order triangular finite elements and piecewise-constant cell conductivities gives
\begin{equation}
\mathbf{A}_j(\boldsymbol{\sigma})
\widehat{\mathbf{u}}_{j,s}
=
\mathbf{b}_{j,s}(\boldsymbol{\sigma}),
\label{eq:discrete_forward}
\end{equation}
with
\begin{equation}
\mathbf{A}_j(\boldsymbol{\sigma})
=
\sum_{e=1}^{N_c}
\sigma_e
\left(
\mathbf{K}_e+k_j^2\mathbf{M}_e
\right)
+
\mathbf{B}_j(\boldsymbol{\sigma}),
\label{eq:system_matrix}
\end{equation}
where $\mathbf{K}_e$ and $\mathbf{M}_e$ are the element stiffness and mass matrices, and $\mathbf{B}_j$ contains the boundary contributions. The possible dependence of $\mathbf{b}_{j,s}$ on conductivity is retained because the forward problem is solved using a primary--secondary formulation.

The potential at the survey plane is reconstructed by numerical quadrature over wavenumber,
\begin{equation}
\mathbf{u}_s(y=0)
=
\frac{1}{\pi}
\int_0^\infty
\widehat{\mathbf{u}}_{k,s}\,dk
\approx
\sum_{j=1}^{N_k}
\omega_j\widehat{\mathbf{u}}_{j,s},
\label{eq:inverse_transform}
\end{equation}
where $\omega_j$ includes the normalization of the cosine transform.

For a quadrupole $q=(A_q,B_q,M_q,N_q)$, the transfer resistance is
\begin{equation}
R_q
=
\left(
\mathbf{p}_{M_q}-\mathbf{p}_{N_q}
\right)^T
\sum_{j=1}^{N_k}
\omega_j
\left(
\widehat{\mathbf{u}}_{j,A_q}
-
\widehat{\mathbf{u}}_{j,B_q}
\right),
\label{eq:transfer_resistance}
\end{equation}
and the apparent resistivity is
\begin{equation}
\rho_{a,q}=|G_qR_q|,
\label{eq:apparent_resistivity}
\end{equation}
where $\mathbf{p}_e$ is the interpolation vector at electrode $e$ and $G_q$ is the geometric factor.

The finite-element solve is incorporated into automatic differentiation through implicit differentiation rather than by differentiating the operations of the sparse factorization. For each wavenumber and source, define the discrete residual
\begin{equation}
\mathbf{r}_{j,s}
\left(
\widehat{\mathbf{u}}_{j,s},
\boldsymbol{\sigma}
\right)
=
\mathbf{A}_j(\boldsymbol{\sigma})
\widehat{\mathbf{u}}_{j,s}
-
\mathbf{b}_{j,s}(\boldsymbol{\sigma})
=
\mathbf{0}.
\label{eq:discrete_residual}
\end{equation}
Because
$\partial\mathbf{r}_{j,s}/\partial\widehat{\mathbf{u}}_{j,s}
=\mathbf{A}_j$,
the implicit function theorem gives the directional derivative of the field for a perturbation $\delta\boldsymbol{\sigma}$:
\begin{equation}
\mathbf{A}_j
\delta\widehat{\mathbf{u}}_{j,s}
=
\delta\mathbf{b}_{j,s}
-
\delta\mathbf{A}_j
\widehat{\mathbf{u}}_{j,s}.
\label{eq:implicit_jvp}
\end{equation}
Equation~\eqref{eq:implicit_jvp} defines the Jacobian--vector product required for forward sensitivity calculations without forming the derivative of the sparse solver itself.

For reverse-mode automatic differentiation, let
\begin{equation}
\overline{\mathbf{u}}_{j,s}
=
\frac{\partial\Phi}
{\partial\widehat{\mathbf{u}}_{j,s}}
\end{equation}
denote the cotangent propagated from the observation operator and the data-misfit term. The corresponding adjoint variable is obtained from
\begin{equation}
\mathbf{A}_j^T\boldsymbol{\lambda}_{j,s}
=
\overline{\mathbf{u}}_{j,s}.
\label{eq:implicit_adjoint}
\end{equation}
The vector--Jacobian product contributed by the PDE solve is then
\begin{equation}
\left.
\frac{\partial\Phi}{\partial\sigma_e}
\right|_{\mathrm{PDE}}
=
\sum_{j,s}
\boldsymbol{\lambda}_{j,s}^T
\left[
\frac{\partial\mathbf{b}_{j,s}}{\partial\sigma_e}
-
\frac{\partial\mathbf{A}_j}{\partial\sigma_e}
\widehat{\mathbf{u}}_{j,s}
\right].
\label{eq:implicit_vjp}
\end{equation}
Any explicit dependence of the objective function on $\boldsymbol{\sigma}$ is added separately by automatic differentiation. If the source vector is independent of conductivity and the Robin coefficient is held fixed in the derivative, equation~\eqref{eq:implicit_vjp} reduces to
\begin{equation}
\left.
\frac{\partial\Phi}{\partial\sigma_e}
\right|_{\mathrm{PDE}}
=
-
\sum_{j,s}
\boldsymbol{\lambda}_{j,s}^T
\left(
\mathbf{K}_e+k_j^2\mathbf{M}_e
\right)
\widehat{\mathbf{u}}_{j,s}.
\label{eq:implicit_vjp_reduced}
\end{equation}

Equations~\eqref{eq:implicit_jvp} and \eqref{eq:implicit_vjp} define the forward- and reverse-mode derivative rules for the ERT solve. In the forward pass, the sparse system in equation~\eqref{eq:discrete_forward} is solved for the electric potential. In the reverse pass, equation~\eqref{eq:implicit_adjoint} is solved using the cotangent supplied by automatic differentiation. The internal factorization steps therefore remain outside the computational graph, while gradients are propagated exactly through the solution of the discrete governing equation.

The derivative with respect to the inversion variable follows from the local model transformation. For example, if $m_e=\log\rho_e$ and $\sigma_e=\exp(-m_e)$, then
\begin{equation}
\frac{\partial\log\rho_{a,q}}{\partial m_e}
=
-\frac{\sigma_e}{R_q}
\frac{\partial R_q}{\partial\sigma_e},
\qquad R_q\neq 0.
\label{eq:log_resistivity_chain}
\end{equation}
The same implicit derivative of the ERT solve can therefore be reused when conductivity is parameterized by resistivity, saturation, water content, or another differentiable model transformation.

The forward and adjoint systems are solved in batches on the GPU. Geometry-dependent quantities and matrix sparsity patterns are reused across wavenumbers, sources, and survey times, whereas the numerical matrix values are updated when the conductivity model changes.

\subsection{Differentiable petrophysical parameterization}

The inversion variable can represent a physical quantity other than electrical resistivity when a differentiable petrophysical relationship is available. Here, water saturation $S$ is related to bulk conductivity through an Archie-type relationship with a surface-conduction term \citep{Archie1942,Waxman1968}
\begin{equation}
\sigma(S)
=
\sigma_pS^n+\sigma_sS^{n-1},
\label{eq:archie_surface}
\end{equation}
where $\sigma_p$ represents the pore-fluid contribution, $\sigma_s$ represents the surface-conduction contribution, and $n$ is the saturation exponent. Volumetric water content is then
\begin{equation}
\theta=\phi S,
\label{eq:water_content}
\end{equation}
where $\phi$ is porosity.

To enforce the saturation bounds, an unconstrained variable $m_e$ is mapped to saturation using
\begin{equation}
s(m_e)=\frac{1}{1+\exp(-m_e)},
\qquad
S(m_e)
=
S_{\min}
+
(1-S_{\min})s(m_e).
\label{eq:saturation_mapping}
\end{equation}
The inverse problem is therefore written as
\begin{equation}
\begin{aligned}
\Phi(\mathbf{m})
={}&
\Phi_d
\left[
\mathcal{F}_{\sigma}
\left(
\boldsymbol{\sigma}
\left(
\mathbf{S}(\mathbf{m})
\right)
\right),
\mathbf{d}_{\mathrm{obs}}
\right]
\\
&+
\alpha_s
\Phi_s
\left(
\boldsymbol{\theta}(\mathbf{m})
\right)
+
\alpha_t
\Phi_t
\left(
\boldsymbol{\theta}(\mathbf{m})
\right)
+
\alpha_c
\Phi_c
\left(
\boldsymbol{\theta}(\mathbf{m})
\right).
\end{aligned}
\label{eq:petrophysical_objective}
\end{equation}
The local derivative passed to the ERT forward model is
\begin{equation}
\frac{\partial\sigma_e}{\partial m_e}
=
\left[
n\sigma_pS_e^{n-1}
+
(n-1)\sigma_sS_e^{n-2}
\right]
(1-S_{\min})
s(m_e)
\left[
1-s(m_e)
\right].
\label{eq:petrophysical_derivative}
\end{equation}
Automatic differentiation propagates this derivative through the forward model and the regularization terms. The recovered water-content field is therefore a petrophysically constrained estimate rather than a direct measurement by ERT.

\section {Synthetic examples and verification}
\subsection{Synthetic model}
\begin{figure}
\centering
\includegraphics[width=\textwidth]{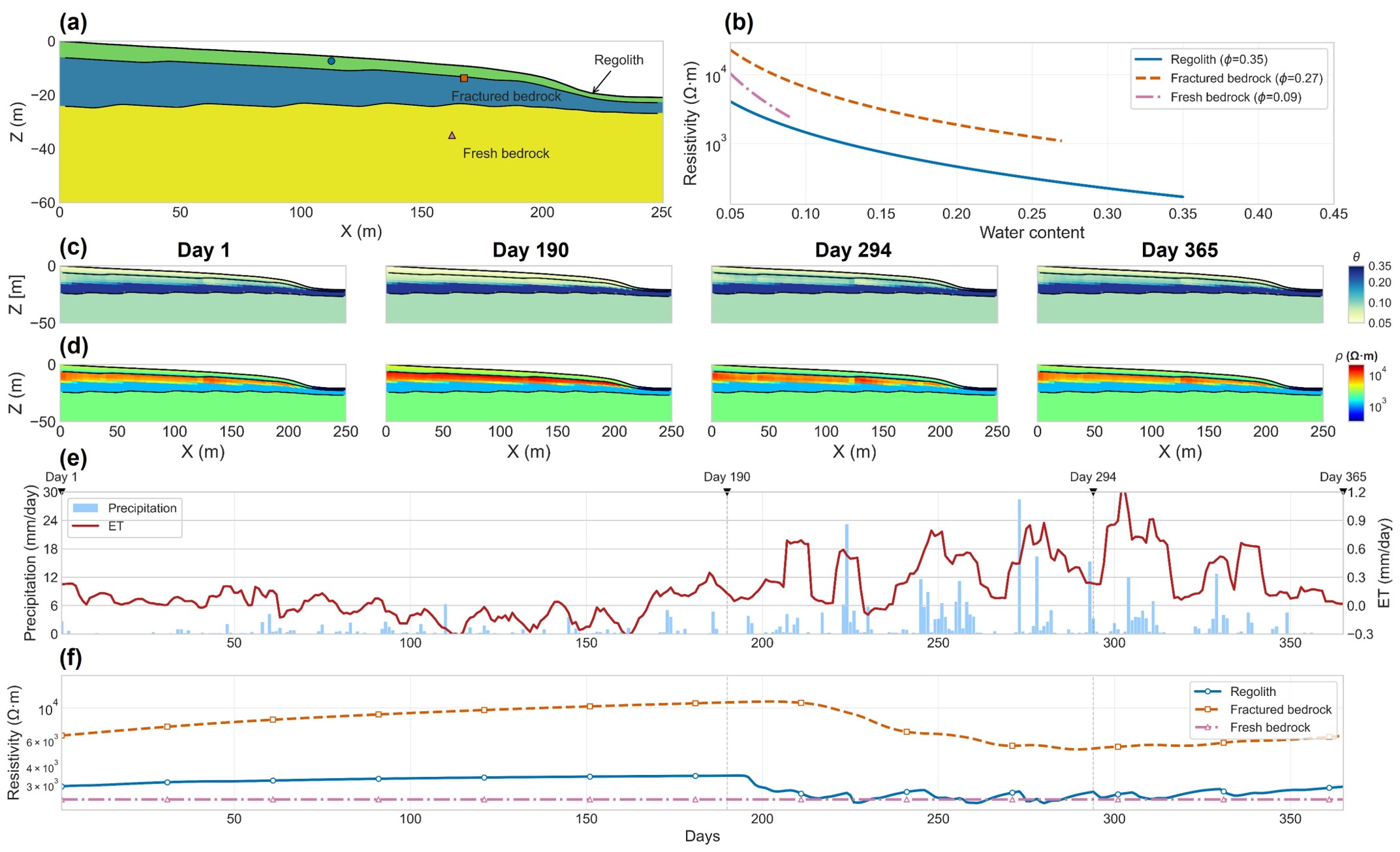}
\caption{Construction of the synthetic time-lapse resistivity models from a hillslope hydrologic simulation. 
(a) ParFlow--CLM hillslope model with regolith, fractured bedrock, and fresh bedrock. 
(b) Relationships between volumetric water content and resistivity. 
(c) Simulated water-content fields at day 1, day 190, day 294, and day 365. 
(d) Resistivity models derived from the water-content fields. 
(e) Precipitation and evapotranspiration data for the one-year simulation; dashed lines mark the four selected times. 
(f) Resistivity time series for the regolith, fractured-bedrock, and fresh-bedrock units.}
\label{fig2}
\end{figure}

We constructed a synthetic time-lapse model from a hillslope-scale ParFlow--CLM simulation \citep{Kollet2008,Maxwell2015}. The model represents a 250-m-long transect in the Laramie Range, Wyoming, near 41.184$^{\circ}$N, 105.396$^{\circ}$W. The subsurface critical zone was represented by three hydrostratigraphic units: regolith, fractured bedrock, and fresh bedrock (Fig.~\ref{fig2}a). Meteorological forcing was taken from the NLDAS-2 data set \citep{Xia2012} for water year 2023, from 1 October 2022 to 30 September 2023, at hourly resolution (Fig.~\ref{fig2}e). The ParFlow--CLM simulation provided the water-content fields used to construct the synthetic time-lapse ERT models. These water-content fields were converted to resistivity using prescribed petrophysical relationships with PyHydroGeophysX \citep{Chen2026PyHydroGeophysX} defined separately for the three hydrostratigraphic units, following the parameterization of \citet{ChenNiu2022} (Fig.~\ref{fig2}b). Four representative times were selected for visualization and inversion tests: day 1, day 190, day 294, and day 365. These snapshots represent the initial condition, the onset of stronger temporal variability, a wet period, and the final state of the simulation.  {The mapping from water content to resistivity produced a sequence of resistivity models representing temporal changes driven by the simulated hydrologic response} (Fig.~\ref{fig2}c and d).

During the first half of the simulation, precipitation was limited, and the sampled regolith and fractured bedrock points show a gradual increase in resistivity, indicating a drying tendency in the shallow and intermediate parts of the model. After day 190, more frequent precipitation events produced a rapid decrease and stronger short-term fluctuations in the regolith resistivity. The fractured bedrock point also shows a decrease in resistivity after the wetting period begins, but its response is smoother and more delayed than that of the regolith. In contrast, the fresh-bedrock point remains nearly unchanged throughout the simulation. This weak response is consistent with its lower porosity and more compact texture, which limit water storage and short-term hydrologic variability in the fresh bedrock unit. These contrasts provide a synthetic time-lapse ERT data set with rapid shallow changes, delayed intermediate responses, and a nearly stable deep background.

\begin{figure}
\centering
\includegraphics[width=\textwidth]{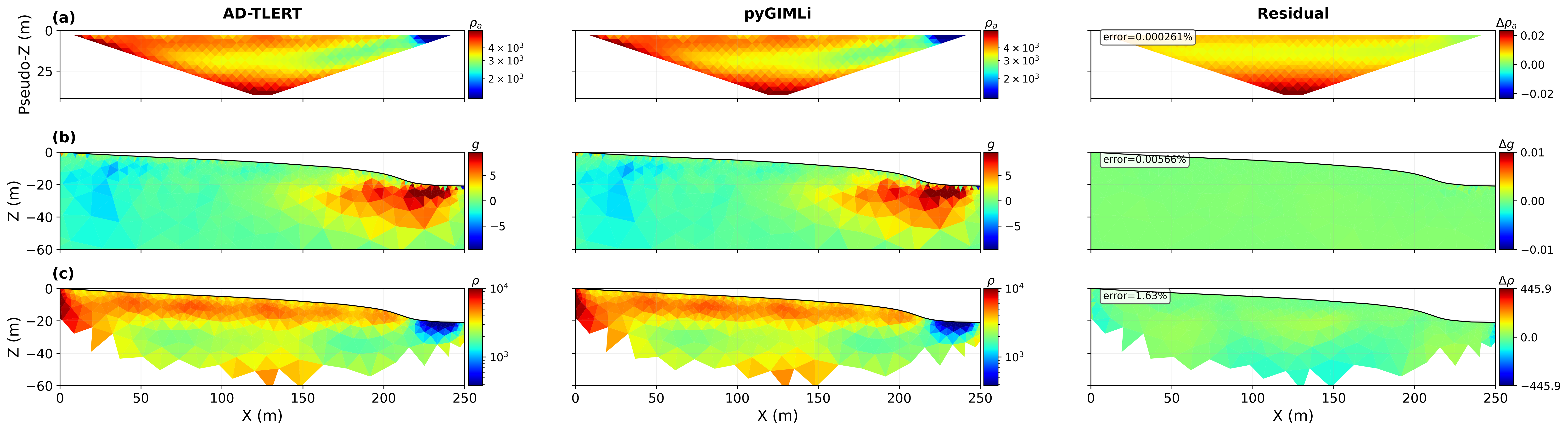}
\caption{Numerical verification of AD-TLERT against pyGIMLi. (a) Apparent resistivity pseudosections from AD-TLERT and pyGIMLi and  {their residual on day 190}. (b) Gradients from the two implementations and their residual. (c) Recovered resistivity models and their residual.}
\label{fig3}
\end{figure}

\subsection{Numerical verification and computational performance}
This experiment, on the synthetic hillslope model, evaluates the two central claims of AD-TLERT: the accuracy of the reusable derivative implementation and the computational efficiency of the GPU workflow. We therefore compare apparent resistivities, gradients, recovered models, and elapsed time with pyGIMLi, which served as the reference ERT solver \citep{Rucker2017}. A total of 48 electrodes were uniformly distributed along the topographic surface. The time-lapse inversions were carried out using a moving window strategy with a window length of three surveys \citep{Wilkinson2022}. This strategy couples neighboring surveys through local temporal constraints, avoiding the computational cost of jointly inverting the entire monitoring sequence. All numerical experiments were conducted on a local workstation equipped with an Intel Core i7-14700KF CPU and an NVIDIA GeForce RTX 4070 GPU.

The two forward responses are nearly indistinguishable over the full pseudosection (Fig.~\ref{fig3}a). The residual is small and shows no systematic pattern associated with electrode spacing or topography. The relative error between the two apparent resistivity responses is \(2.61 \times 10^{-4}\%\), indicating that the differentiable forward solver reproduces the reference ERT response for the tested terrain geometry. We then compared  {the gradient $\mathbf{g}$ computed by AD-TLERT} with the corresponding reference result from pyGIMLi (Fig.~\ref{fig3}b). The two gradient fields have the same spatial pattern, with the largest amplitudes concentrated in the downslope part of the section. The residual is close to zero over most of the model domain, and the relative difference is \(5.66 \times 10^{-3}\%\). This agreement verifies that the reverse-mode automatic differentiation correctly propagates the data misfit through the ERT forward calculation to the model parameters.

Finally, we compared the recovered resistivity models from the two workflows (Fig.~\ref{fig3}c). Both inversions recover the same resistivity structure, including the high-resistivity shallow region along the upper slope and the lower-resistivity region near the downslope end of the profile. The relative difference between the recovered resistivity models is \(1.63\%\), which is small relative to the total resistivity contrast. In addition to the numerical agreement, AD-TLERT provided a substantial reduction in the elapsed time. Under the tested configuration, the AD-TLERT workflow completed the inversion in 8.66 min,  {whereas the pyGIMLi-based workflow required 441.95 min, corresponding to an approximately 51.0-fold speedup.}

\section{Effects of inversion choices}

The AD-TLERT framework allows the main components of the inverse problem to be modified within the same differentiable workflow and modular structure. Therefore, it is convenient to examine  {how representative choices of data-misfit function, regularization strategy, and optimization method affect the recovered resistivity.} In this section, we vary one component of the workflow at a time  {while keeping the other parameters and the windowed time-lapse strategy fixed.}

\begin{figure}
\centering
\includegraphics[width=\textwidth]{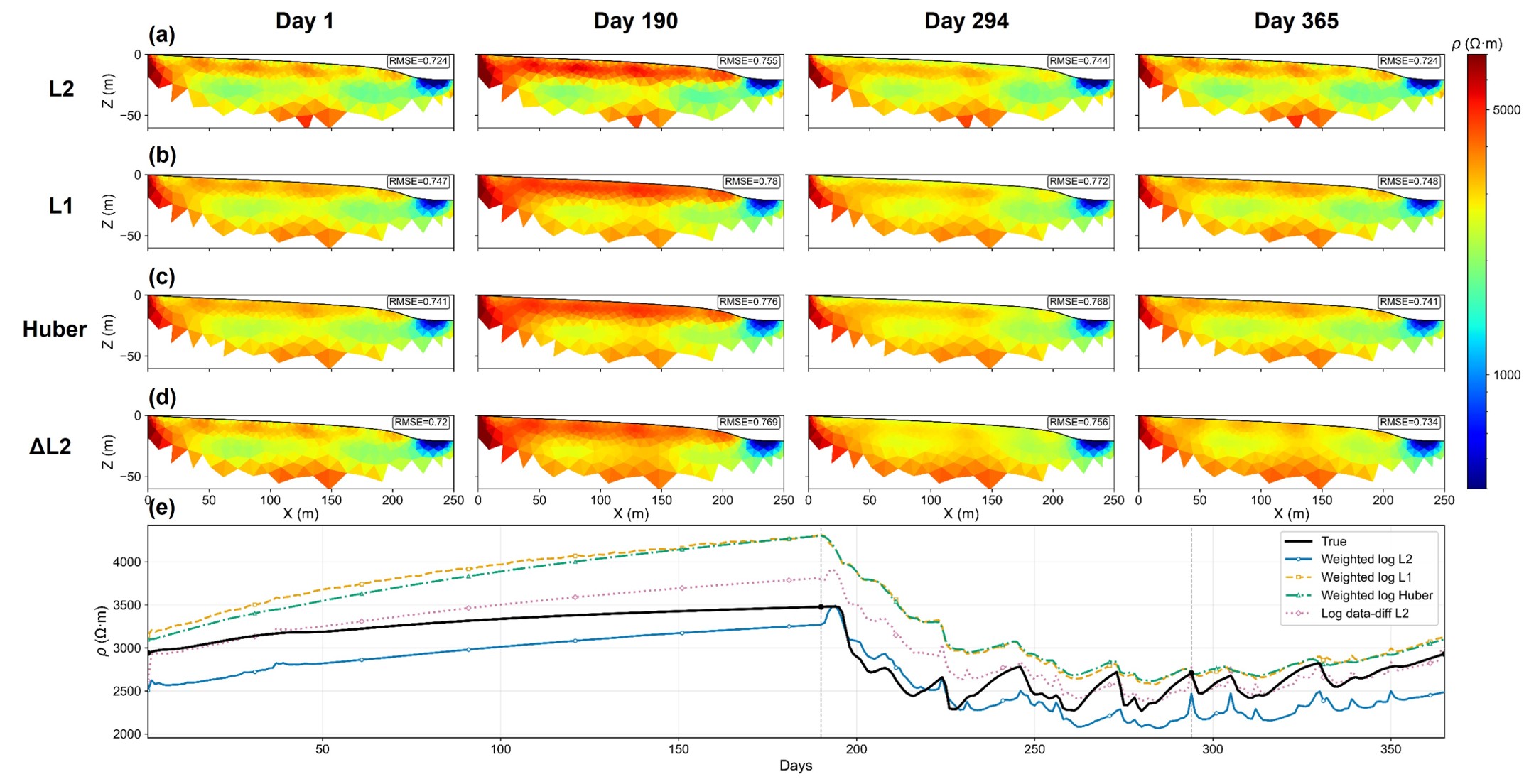}
\caption{Effect of data misfit functions on synthetic AD-TLERT inversion. Recovered resistivity models at day 1, day 190, day 294, and day 365 using (a) L2, (b) L1, (c) Huber, and (d) difference L2 misfit functions. (e) Resistivity time series extracted from the representative point in regolith, comparing the true model with the results from the four data misfits.}
\label{fig4}
\end{figure}

\subsection{Effect of data misfit functions}
We first evaluated how the data misfit function affects the recovered time-lapse resistivity models. Four representative formulations were tested: L2, L1, Huber, and data-difference L2. These choices represent different assumptions about the residuals and about the quantity that should be fitted.  {The L2 misfit provides the standard least-squares baseline and places disproportionately greater weight on large residuals.} The L1 and Huber misfits were included as robust alternatives that reduce the influence of large residuals \citep{Huber1964}, which can arise in field ERT data from measurement noise, changing contact conditions, or imperfect error estimates. The data-difference L2 formulation was included to represent a time-lapse objective that emphasizes changes relative to a reference survey rather than fitting each survey only in an absolute sense. The first three formulations fit the apparent-resistivity data at each survey, whereas the data-difference formulation fits the change in apparent resistivity relative to the reference state.

All four formulations recover the main structure of the synthetic model (Fig.~\ref{fig4}). The high-resistivity region along the upper slope and the lower-resistivity zone near the downslope end are present in all cases,  {indicating that the large-scale structure is not controlled by the choice of misfit alone.} The differences are instead expressed mainly in model amplitude and temporal evolution. The L2 inversion gives the lowest or nearly lowest RMSE for most selected snapshots, with values of 0.724, 0.755, 0.744, and 0.724 for day 1, day 190, day 294, and day 365, respectively. This behavior is expected for a controlled synthetic data set without strong outliers, because the L2 objective uses the full residual magnitude during the update. The L1 and Huber inversions produce similar spatial patterns but recover slightly higher resistivity in the shallow and intermediate parts of the section, which increases the RMSE at the selected times. In this example, their reduced sensitivity to large residuals makes the update more conservative, so the robust objectives do not improve the clean synthetic case in terms of model RMSE.

The representative time series highlights these differences more clearly than the model snapshots alone (Fig.~\ref{fig4}e). Before the main wetting period, the L1 and Huber results overestimate the gradual resistivity increase, whereas the L2 result remains below the true resistivity level. After approximately day 190, all methods capture the rapid resistivity decrease caused by the increase in water content, but they recover different amplitudes. The L2 result follows the timing of the drop but underestimates the post-wetting resistivity level. The L1 and Huber results remain higher over part of the wet period, consistent with their more conservative response to large residuals. The data-difference L2 inversion follows the main timing of the resistivity decrease and gives a smoother temporal response, although it does not yield the lowest RMSE for every selected snapshot. This behavior reflects the purpose of the data-difference objective: it prioritizes relative changes from the reference state and is less focused on matching the absolute resistivity level at each survey.

These results show that the data misfit function affects different aspects of the time-lapse inversion. In this controlled synthetic example, the L2 objective provides the strongest baseline for recovering the absolute resistivity model. The robust objectives are less favorable in terms of RMSE, but they are useful choices for field data where large residuals may otherwise dominate the inversion. The data-difference objective is most relevant when the interpretation focuses on temporal changes rather than the absolute model at each survey.

\begin{figure}
\includegraphics[width=\textwidth]{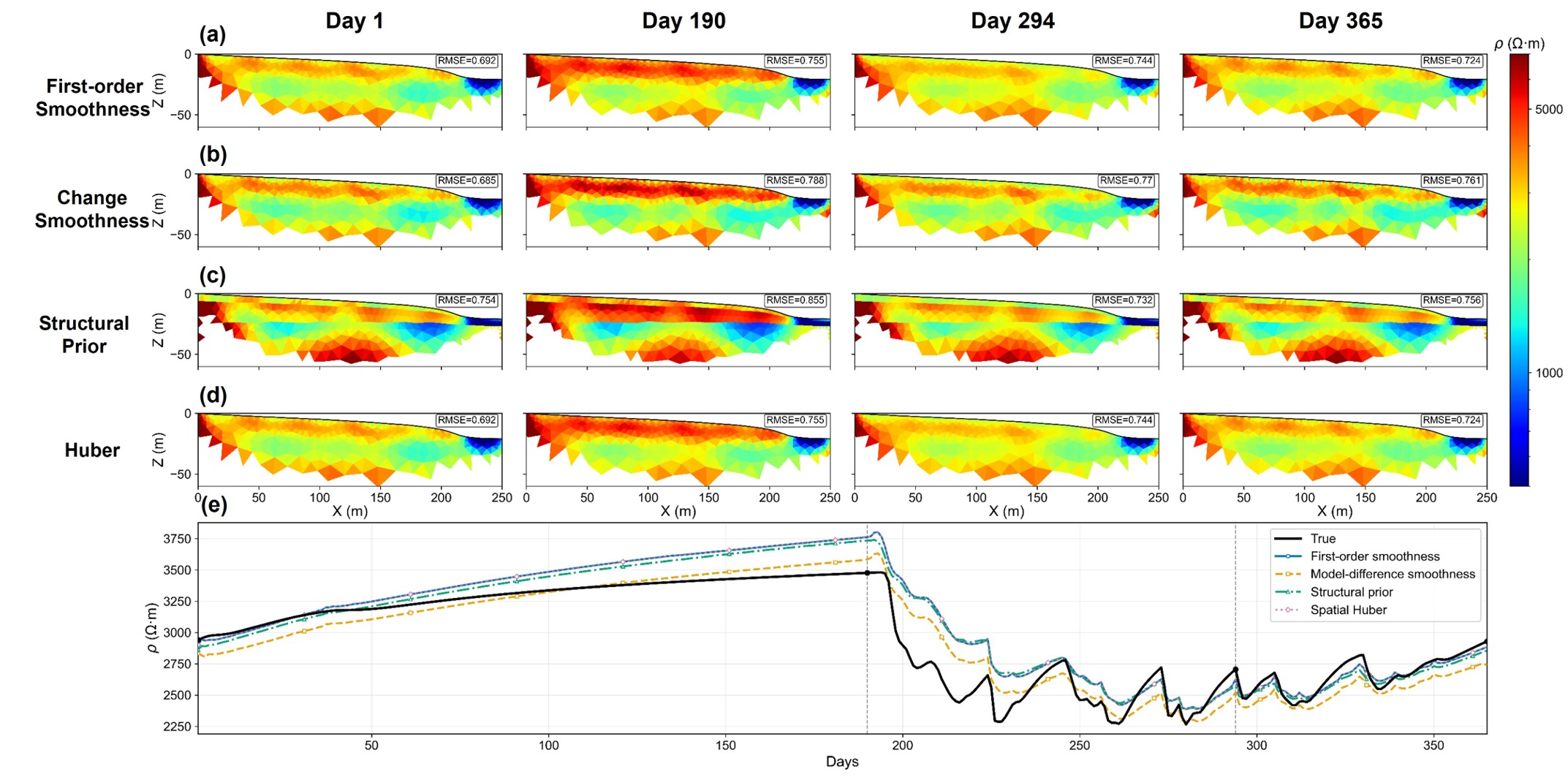}
\caption{Effect of spatial regularization on synthetic AD-TLERT inversion. Recovered resistivity models at day 1, day 190, day 294, and day 365 using (a) first-order smoothness,  {(b) model-difference smoothness,} (c) structural prior regularization, and (d) spatial Huber regularization. (e) Representative resistivity time series for the true model and the four regularization results.}
\label{fig5}
\end{figure}

\subsection{Effect of regularization}

 {Four representative spatial regularization strategies were compared:}  {first-order smoothness, model-difference smoothness, structural prior regularization}, and spatial Huber regularization. These choices were selected because they impose different assumptions on the model. First-order smoothness provides the standard baseline by penalizing spatial roughness in the recovered resistivity model.  {Model-difference smoothness regularizes} the spatial variation of changes relative to a reference model, which is more closely aligned with the time-lapse objective of recovering changes. The structural prior uses the prescribed hydrostratigraphic structure to guide the spatial constraint, whereas the Huber regularization provides an edge-preserving alternative that behaves like a smoothness constraint for small gradients but reduces the penalty on larger contrasts.

 {The first-order smoothness result (Fig.~\ref{fig5}a) provides a stable baseline, with RMSE values of 0.692, 0.755, 0.744, and 0.724 for days 1, 190, 294, and 365, respectively.}  {The model-difference smoothness result (Fig.~\ref{fig5}b)} gives a slightly lower RMSE at the initial snapshot but larger errors at later times. This behavior suggests that constraining the changes can preserve the reference structure, but it may also limit the recovery of larger time-dependent amplitudes when the true change becomes more pronounced. The structural prior result (Fig.~\ref{fig5}c) departs more strongly from the other regularization results. In particular, the localized low-resistivity anomaly near the downslope end  {is not fully recovered at its true amplitude.}  {This anomaly is difficult to reconstruct because it is localized and lies close to the end of the profile, where sensitivity and spatial resolution are limited.} Under the structural constraint, the inversion has less freedom to place the remaining change entirely within this localized low-resistivity zone. Part of the unresolved data misfit  {appears to be compensated for by changes elsewhere in the model, such as in the fresh bedrock.} This behavior is especially evident at day 190. The result does not imply that structural information is generally unfavorable; rather, it shows that regularization from prior information must be weighted carefully. The spatial Huber result is close to the first-order smoothness result, but its role is different. It provides a compromise between smooth recovery and edge preservation by reducing the influence of large model gradients in the regularization term. In this example, the Huber constraint does not substantially change the large-scale recovered structure compared with first-order smoothness, which is expected because the synthetic model contains broad layer variations rather than isolated sharp anomalies.  {Its main value therefore lies in providing a more flexible constraint for cases where localized changes or sharper wetting fronts are expected.}

The representative time series in Fig.~\ref{fig5}e further shows that the regularization choice affects the temporal amplitude of the recovered model. The first-order smoothness and Huber results follow the general trend of the true curve  {but smooth out part of the temporal variability.}  {The model-difference smoothness result tends to produce} a lower resistivity level before the main wetting period and a more muted response afterward. The structural prior follows the broad changes but introduces amplitude differences associated with the imposed spatial structure. These results show that spatial regularization influences not only the appearance of  {individual resistivity sections but also the recovered temporal evolution} at representative locations.

Overall, the comparison indicates that no single spatial regularization is optimal for all aspects of the time-lapse recovery. First-order smoothness provides a stable reference choice, model-difference smoothness emphasizes changes relative to the baseline, structural prior regularization can preserve known hydrostratigraphic contrasts, and  {Huber regularization provides an edge-preserving alternative.}

\begin{figure}
\centering
\includegraphics[width=\textwidth]{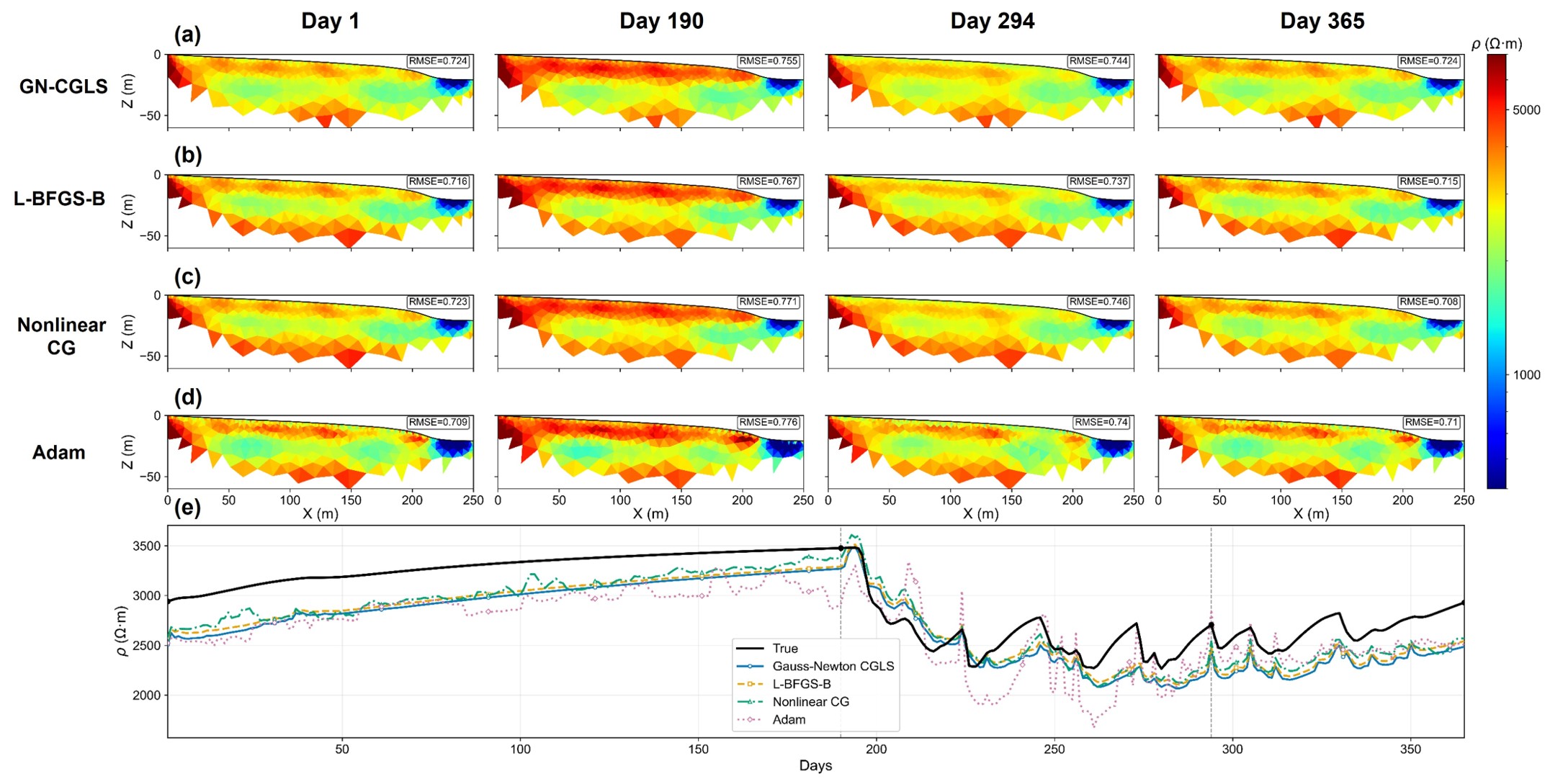}
\caption{Effect of optimization strategy on synthetic AD-TLERT inversion. Recovered resistivity models at day 1, day 190, day 294, and day 365  {using (a) Gauss--Newton CGLS,} (b) L-BFGS-B, (c) nonlinear conjugate gradient, and (d) Adam. (e) Representative resistivity time series for the true model and the four optimizer results.}
\label{fig6}

\vspace{6pt}

\captionof{table}{Comparison of the optimization methods.}
\label{tab:optimizer_comparison}
\small
\begin{tabular*}{\textwidth}
{@{\extracolsep{\fill}}lcccc@{}}
\hline
Optimizer
& Elapsed time (min)
& Peak memory (GB)
& Iterations
& Final objective \\
\hline
Gauss--Newton CGLS & 7.72 & 2.18 & 1904 & 3.02 \\
L-BFGS-B           & 17.33 & 2.12 & 5445 & 3.91 \\
Nonlinear CG       & 22.89 & 2.12 & 5330 & 4.29 \\
Adam               & 31.56 & 2.15 & 5445 & 5.07 \\
\hline
\end{tabular*}
\end{figure}

\subsection{Effect of optimization}
Four optimizers were tested: Gauss--Newton CGLS, L-BFGS-B, nonlinear conjugate gradient (CG), and Adam. They span curvature-informed and first-order update strategies. Gauss--Newton CGLS follows conventional ERT inversion practice by solving a locally linearized least-squares problem. L-BFGS-B constructs a limited-memory approximation to the inverse Hessian and supports bound constraints \citep{Byrd1995}. Nonlinear CG forms search directions from successive gradients, whereas Adam uses adaptive first-order moment estimates \citep{KingmaBa2015}. Their recovered models and representative time series are compared in Fig.~\ref{fig6}, and their computational performance is summarized in Table~\ref{tab:optimizer_comparison}.

All four methods recover the main synthetic resistivity structure, including the resistive upper slope and the conductive downslope zone (Fig.~\ref{fig6}a-d). Differences in snapshot RMSE are modest, and no optimizer gives the lowest RMSE at every selected time.  {The larger differences appear in the temporal response} (Fig.~\ref{fig6}e). Gauss--Newton CGLS, L-BFGS-B, and nonlinear CG recover similar long-term trends, although all underestimate parts of the true resistivity variation. Adam produces stronger short-term oscillations after the main wetting period and departs more frequently from the true curve. The optimizer therefore has a limited effect on the broad spatial pattern in this controlled example but a clearer effect on temporal stability.

Gauss--Newton CGLS is the most computationally efficient method in this comparison. It completes the inversion in 7.72 min with 1904 iterations and reaches the lowest final objective value of 3.02. Its 2.18 GB peak memory use is only slightly higher than that of the other methods. The result reflects the suitability of the Gauss--Newton approximation for a smooth deterministic least-squares problem: local curvature information accelerates objective reduction, while CGLS applies the linearized system through matrix-vector products without storing a full Hessian.

L-BFGS-B reaches a final objective value of 3.91 in 17.33 min, followed by nonlinear CG with 4.29 in 22.89 min. Their recovered sections remain broadly similar to the Gauss--Newton CGLS result (Fig.~\ref{fig6}b,c), but both require substantially more iterations. L-BFGS-B retains practical value when bound constraints or nonlinear parameter transformations are required, because it approximates curvature without a problem-specific Gauss--Newton system. Nonlinear CG has a lower per-iteration complexity but converges less efficiently for this nonlinear and ill-conditioned inverse problem.

Adam has the longest runtime, 31.56 min, and the highest final objective value, 5.07. Its peak memory use remains comparable at 2.15 GB, indicating that memory is not the main source of the performance difference. Instead, its first-order adaptive updates are less effective than curvature-informed directions for this deterministic full-batch inversion. Taken together, Fig.~\ref{fig6} and Table~\ref{tab:optimizer_comparison} show that optimizer selection affects convergence and temporal stability more strongly than the large-scale recovered structure. Gauss--Newton CGLS is the preferred method for the tested resistivity inversion, whereas L-BFGS-B provides a useful alternative when the parameterization requires explicit bounds or greater optimization flexibility.

\begin{figure}
\centering
\includegraphics[width=\textwidth]{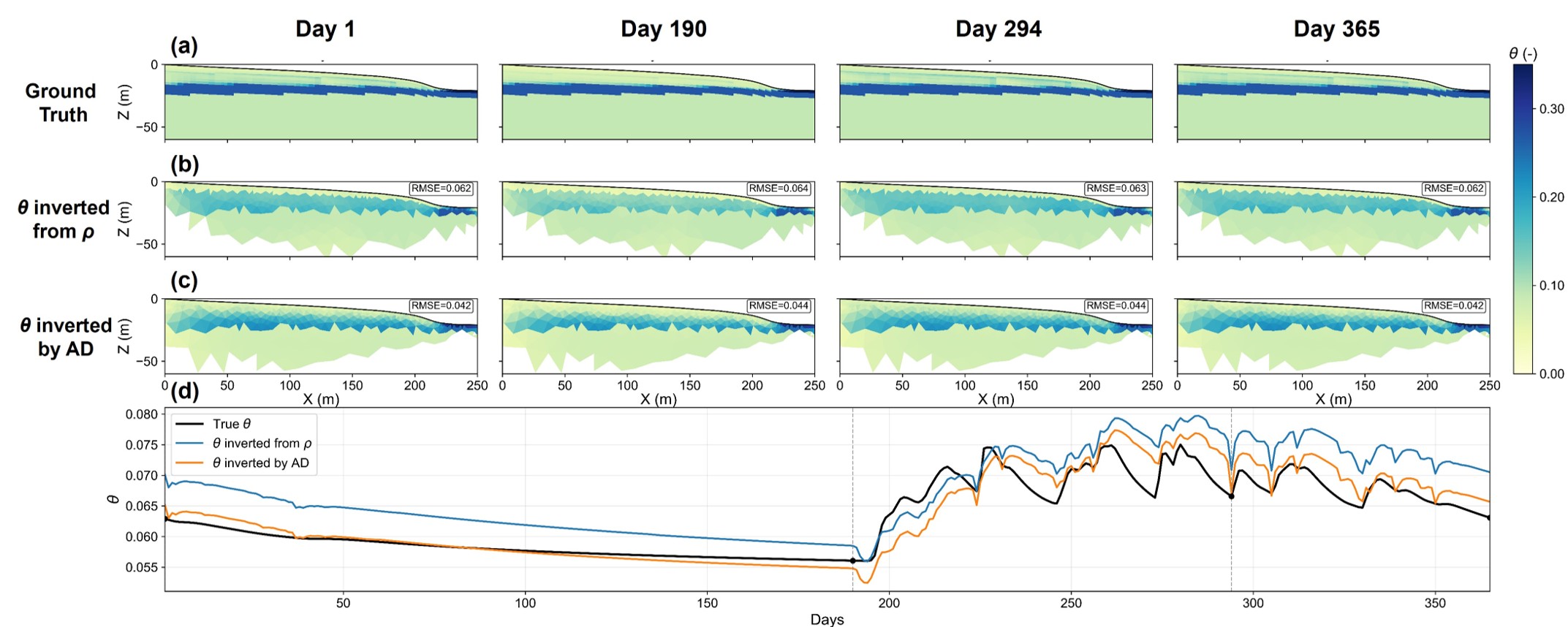}
\caption{Comparison of conversion and direct AD-TLERT water-content inversion. (a) True water-content models at day 1, day 190, day 294, and day 365. (b) Water-content models obtained by converting the recovered resistivity models after inversion. (c) Water-content models recovered by direct AD-TLERT water-content inversion. (d) Representative water-content time series in the regolith.}
\label{fig7}
\end{figure}

\section{Differentiable petrophysical parameterization}

In this experiment, the target parameter is volumetric water content rather than resistivity. The goal is to test whether AD-TLERT can move beyond resistivity imaging and estimate a process-related parameter directly through a differentiable petrophysical relationship. In the first workflow, the inversion is performed in resistivity space, and the recovered resistivity models are converted to saturation and water content after inversion. This represents the conventional interpretation method. In the second workflow, water content is included in the inversion through a differentiable petrophysical parameterization. The optimizer updates the model variable, and the corresponding resistivity model is computed before each ERT forward calculation.

Both workflows recover the main temporal wetting pattern in the shallow and intermediate  {parts of the model (Fig.~\ref{fig7}b,c)}, indicating that the synthetic ERT data contain information about the hydrologic change.  {However, the direct AD-TLERT inversion gives a more accurate water-content estimate.} Across the four representative times, the RMSE values are reduced to 0.042, 0.044, 0.044, and 0.042, corresponding to an average reduction of about \(31\%\) relative to the conversion. The spatial sections show that the conversion result tends to overestimate water content, especially in the shallow to intermediate zone. This bias is also visible in the representative time series, where the converted result remains systematically higher than the true value through most of the simulation.  {The direct AD-TLERT inversion substantially reduces this positive bias.} Although the recovered time series is still smoother than the true response and slightly underestimates some short-term peaks during the wet period, it follows the true curve more closely during both the dry stage and the subsequent wetting period.

These results show that including the petrophysical transformation inside the inversion improves the consistency between the recovered model and the target physical parameter. In the conversion workflow, the inversion is regularized in resistivity space, and the water-content estimate inherits the smoothing and amplitude bias of the recovered resistivity model.  {In the direct AD-TLERT workflow,} the model update is applied to the water-content parameterization itself, while the ERT response is still computed through the corresponding resistivity model. This explains why the direct inversion produces lower RMSE and a more accurate representative time series. The recovered water content should still be interpreted as a petrophysically constrained estimate,  {but it is no longer merely a post-processing product derived from the recovered resistivity model.}

\begin{figure}
\centering
\includegraphics[width=\textwidth]{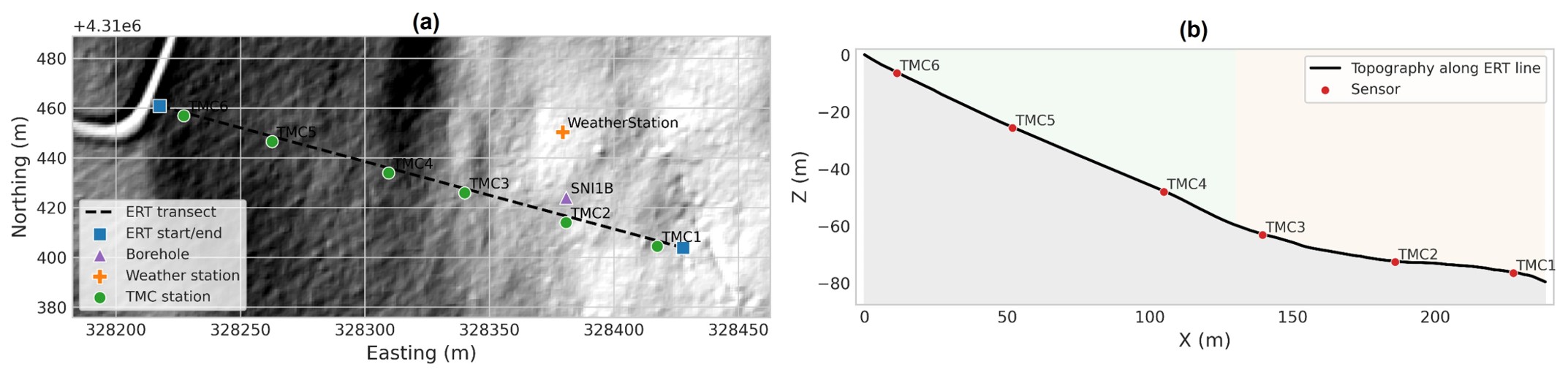}
\caption{Field monitoring setup for the time-lapse ERT experiment. (a) Map view of the ERT profile, TMC soil moisture stations, borehole, and weather station used in the field data set. (b) Topographic profile along the ERT line and projected locations of the monitoring stations.}
\label{fig8}
\end{figure}

\begin{figure}
\centering
\includegraphics[width=\textwidth]{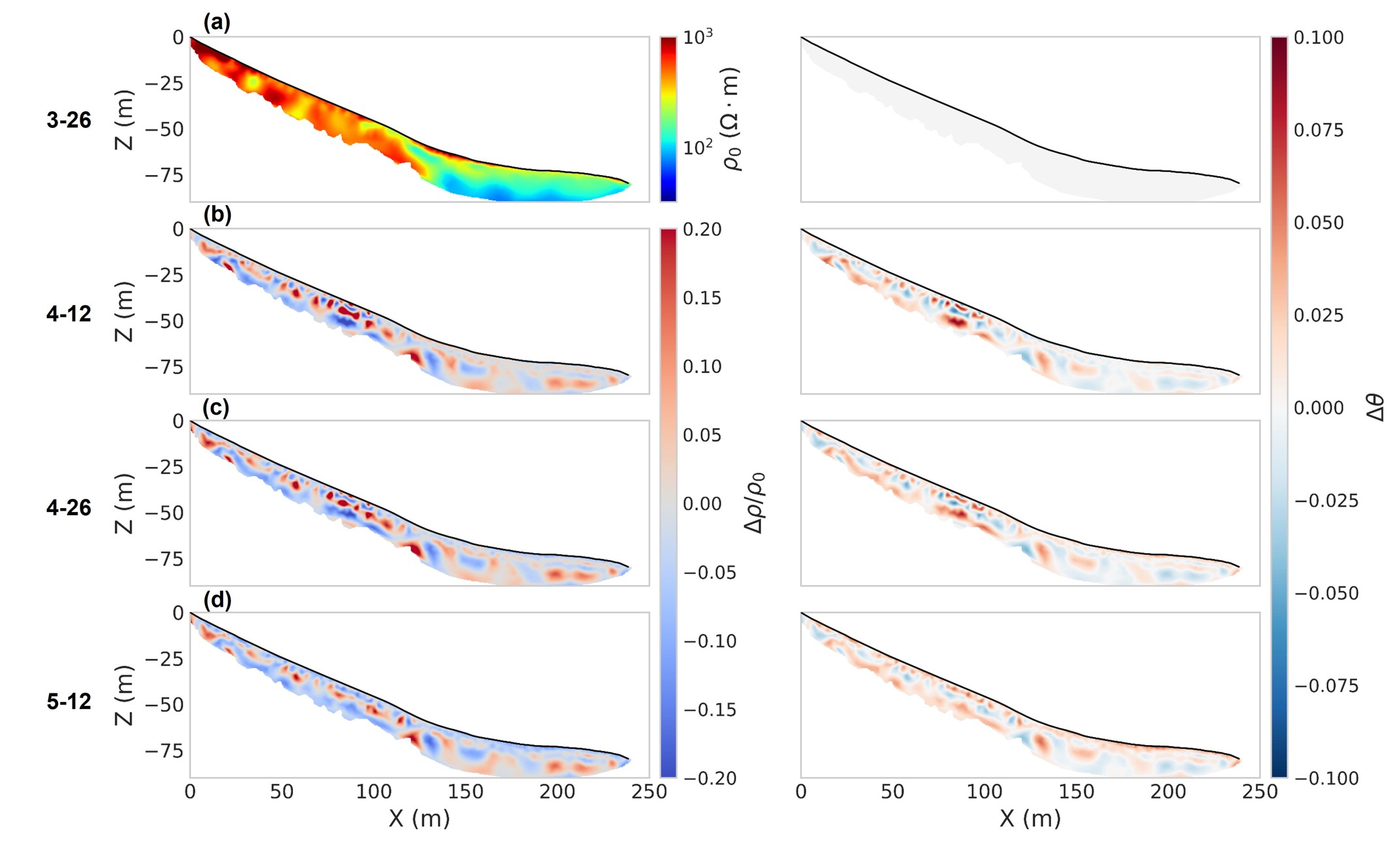}
\caption{The time-lapse inversion results for resistivity and water-content changes during snowmelt. Baseline resistivity model on 26 March 2022 and time-lapse changes on 12 April, 26 April, and 12 May 2022.  {The left column shows the relative resistivity changes with respect to the baseline model, and the right column shows the directly inverted water-content changes obtained with AD-TLERT.}}
\label{fig9}
\end{figure}

\section{Field data inversion}

The field example uses the hillslope monitoring data set reported by \citet{Uhlemann2024}, which includes repeated ERT surveys, TMC soil-moisture sensors, borehole information, and meteorological observations. The ERT transect follows a steep hillslope from the upper slope toward the downslope end, with TMC stations distributed along and near the profile (Fig.~\ref{fig8}a). The topographic profile shows an elevation drop of approximately 80 m over the 250 m transect, and the sensors sample different positions along this gradient (Fig.~\ref{fig8}b). We focus on the period from 26 March to 12 May 2022, using 26 March as the baseline. This interval captures the wetting response from snowmelt and provides a suitable field test for direct water-content-change inversion because both spatially distributed ERT responses and TMC observations are available. In this field example, the temperature dependence of resistivity is accounted for within the inversion using the temperature sensor data, following the correction approach adopted in the original field study. The TMC water-content data are further used as a soft constraint to guide the recovered water-content changes.

\begin{figure}
\centering
\includegraphics[width=\textwidth]{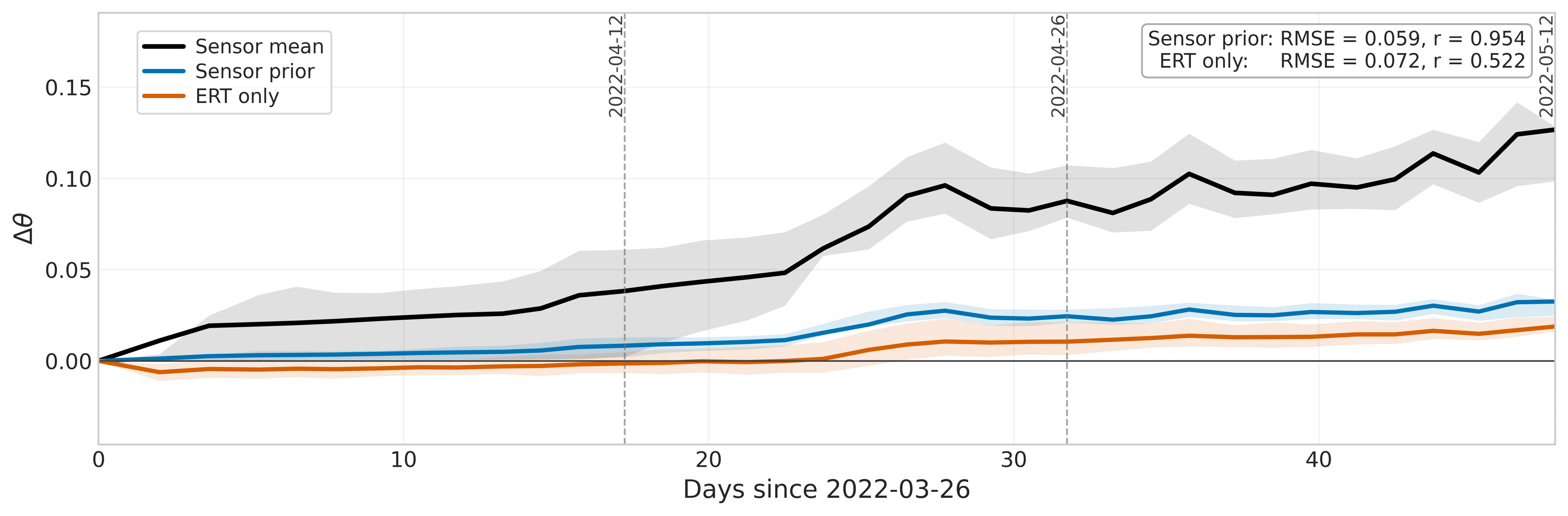}
\caption{Comparison of AD-TLERT water-content changes with TMC sensor observations. Time series of water-content change from TMC sensors, direct AD-TLERT inversion with sensor prior, and ERT-only inversion. The shaded regions indicate the interquartile range of the sensor observations.}
\label{fig10}
\end{figure}

The baseline model on 26 March provides the reference resistivity structure along the hillslope, with higher resistivity in the upper and shallow parts of the profile and a lower-resistivity zone toward the downslope end (Fig.~\ref{fig9}). The subsequent panels show the relative resistivity changes and the directly inverted water-content changes for 12 April, 26 April, and 12 May. Relative to the 26 March baseline, the later surveys show negative resistivity changes mainly in the shallow and intermediate parts of the profile. The response is spatially heterogeneous, which agrees with the field interpretation of \citet{Uhlemann2024} that snowmelt-driven wetting and flow are controlled by hillslope position, bedrock structure, and near-surface storage. Shallow lateral flow dominates the steep upper slope, whereas more vertical or groundwater-related dynamics occur in the lower colluvial part of the transect.

Fig.~\ref{fig10} compares the inferred water-content changes with the TMC data. The ERT inversion captures part of the snowmelt wetting trend, but its agreement with the sensor mean is limited, with an RMSE of 0.072 and a correlation of 0.522. After incorporating the TMC observations as a soft constraint, the recovered water content follows the trend of the sensor response much more closely. The RMSE decreases to 0.059, and the correlation increases to 0.954. This comparison shows that the sensor information improves the temporal consistency of the direct water-content inversion, especially during the main wetting period. The recovered amplitudes remain lower than the sensor mean, which is expected because the TMC probes  {measure local water content at a point,} whereas ERT provides a weighted and regularized estimate over a larger support volume.

\section{DISCUSSION}

\subsection{Methodological flexibility and computational efficiency}

The experiments show that the main value of AD-TLERT lies in separating the PDE derivative from the upper-level inversion formulation while reducing the cost of repeated forward and adjoint calculations in long monitoring sequences. The comparison with pyGIMLi \citep{Rucker2017} confirms the numerical consistency of the forward responses, gradients, and recovered resistivity models. The supplementary comparison with ResIPy, using its R2 inversion backend \citep{Blanchy2020}, provides an additional independent test of model recovery. Although ResIPy uses a substantially denser default inversion mesh and is therefore not used for a direct runtime comparison, both implementations recover similar large-scale resistivity structures and time-lapse changes. ResIPy gives slightly lower errors for the absolute resistivity models, whereas AD-TLERT gives slightly lower errors for the recovered changes at the three representative times (Supplementary Figs~S1 and S2). These results indicate that the main time-lapse responses recovered by AD-TLERT are consistent with those obtained using an established independent inversion code.

The comparison with ADERT \citep{Rincon2026} provides a more direct assessment of computational efficiency because both frameworks use automatic differentiation. On the official ADERT 2.5-D reference model, the predicted data and gradients from the two implementations are highly correlated, while AD-TLERT reduces the combined forward-and-gradient time by a factor of approximately 10.3 and lowers the full-process peak GPU memory from 961 to 485~MiB (Supplementary Table~S4). This improvement is primarily associated with implicit differentiation of the sparse PDE solve, cuDSS sparse direct factorization, batched multiple-right-hand-side solves, and reuse of matrix structures and numerical factors.

The subsequent experiments further show that data-misfit functions, regularization terms, optimization algorithms, petrophysical transformations, and constraints based on auxiliary observations can be evaluated within the same computational framework. AD-TLERT therefore provides a reusable platform for examining how alternative inversion assumptions affect time-lapse ERT results and hydrologic interpretation.

\subsection{Limitations}
First, the forward model is 2.5-D and assumes that conductivity is invariant perpendicular to the survey line. This approximation is appropriate where the dominant electrical variations lie within the profile plane, but it cannot represent off-profile heterogeneity, fully three-dimensional topography, or three-dimensional flow paths. The present GPU implementation has also been evaluated only for 2.5-D problems, and extending the sparse solves to a full three-dimensional domain will impose substantially greater memory and computational demands. Second, the synthetic experiments were designed as controlled tests and used prescribed petrophysical relationships. The improved recovery of water-content therefore demonstrates the benefit of updating and regularizing the target variable when the adopted relationship is adequate, but it does not remove uncertainty in porosity, pore-water conductivity, saturation exponent, surface conduction, or their spatial and temporal variability. Such petrophysical uncertainty can strongly affect water-content estimates derived from field ERT data \citep{Tso2019}. These parameters are treated as known in the present study, and their uncertainty is not propagated into the recovered models.

\subsection{Future work}
Future development will proceed in four related directions. First, extending the forward and adjoint solvers to a fully three-dimensional spatial domain will allow repeated 3-D surveys to be inverted in space and time, yielding a full 4-D formulation for sites where off-profile structure is important \citep{Kim2009,Loke2014}. This extension will require memory-efficient sparse or iterative solvers, matrix-free derivative products, and potentially multi-GPU execution. Second, the differentiable framework provides a natural basis for Bayesian inversion and uncertainty quantification. Gradient-based sampling, variational inference, and ensemble methods could be used to quantify uncertainty arising from data noise, model non-uniqueness, regularization, and petrophysical parameters \citep{Ramirez2005,Oware2019,Khabaz2024,Yan2026}. Third, the grid-based model could be replaced or complemented by an implicit neural representation. A coordinate-based neural network could represent resistivity, water content, or their temporal changes as a continuous function of spatial and temporal coordinates, with its weights estimated directly from the ERT observations through the differentiable forward model. In a test-time formulation, this approach would not require a pretraining data set and could provide a shared space--time representation together with network-induced implicit regularization. Related neural reparameterizations have been demonstrated in seismic full-waveform inversion and direct-current resistivity inversion \citep{Sun2023IFWI,Xu2024TTL,Xu2025NeuralFields}. The effects of network architecture, coordinate encoding, representation bias, and explicit physical constraints will require systematic evaluation. Finally, AD-TLERT can be extended to joint inversion of complementary geophysical observations and to coupled hydrogeophysical inversion. ERT could be combined with GPR or other geophysical data through structural or petrophysical coupling, whereas hydrological process models could be embedded to estimate water content and hydraulic parameters more directly \citep{Linde2006,Hinnell2010,Pleasants2022}. These extensions would follow the same computational principle, provided that each coupled component is differentiable or supplied with an appropriate implicit derivative.

\section{CONCLUSIONS}

We developed AD-TLERT to address two practical limitations of long-sequence time-lapse ERT inversion: the computational cost of repeated forward and adjoint calculations, and the derivative-development effort required when changing the inversion formulation. The framework combines automatic differentiation of the upper-level inversion components with implicit differentiation of the sparse 2.5-D finite-element problem. This separation allows data-misfit functions, regularization terms, model transformations, and constraints based on auxiliary observations to reuse the same PDE derivative implementation. GPU sparse factorization, batched multiple-right-hand-side solves, reuse of geometry-dependent quantities, and windowed temporal coupling further reduce the computational cost. Comparisons with pyGIMLi showed close agreement in apparent resistivities, gradients, and recovered resistivity models. Under the tested hardware and inversion configuration, AD-TLERT reduced the elapsed time from 441.95 to 8.66 min, corresponding to an approximately 51-fold speedup.

Synthetic experiments demonstrated that alternative data-misfit functions, regularization strategies, and optimization algorithms can be evaluated within the same computational framework and that their effects are expressed in the recovered amplitudes, spatial structure, temporal behavior, and computational cost. Differentiable petrophysical parameterization further enabled water content to be estimated directly rather than through post-inversion conversion, reducing the average water-content error by approximately 31\% for the tested synthetic model. The field application showed that temperature information and soil-moisture observations can also be incorporated into the inversion workflow to constrain snowmelt-driven hillslope wetting. AD-TLERT therefore provides a computationally efficient and extensible platform for testing time-lapse inversion assumptions, incorporating complementary observations, and extending ERT from resistivity imaging towards the estimation of process-relevant subsurface variables.

\section*{Acknowledgments}
This work was supported by the University of Iowa Old Gold Summer Fellowship and startup fund from the University of Iowa.

\urlstyle{same}
\section*{Data Availability}
The core source code for AD-TLERT is publicly available at \url{https://github.com/Vcholerae1/AD-TLERT}. The repository also contains the hydrological-model files and corresponding resistivity models used in the synthetic example, together with a compact comparison example involving AD-TLERT, ResIPy, and pyGIMLi. The complete scripts and configuration files required to reproduce all experiments reported in this study will be released in the same
repository upon acceptance. The field ERT, soil-moisture, and temperature data used in the snowmelt application are available from the ESS-DIVE repository \citep{UhlemannDataset2024}.

\bibliographystyle{ad_tlert_references}
\bibliography{adert_references}

\end{document}


\maketitle

This Supplementary Information gives the numerical details and parameter settings used in the AD-TLERT experiments. The notation follows the main text. The first section describes the GPU implementation, Sections~S2 and S3 specify the inversion and synthetic-model settings, Section~S4 documents the cross-code comparisons, and Section~S5 gives the temperature correction and sensor constraint used in the field example.

\clearpage

\section{Numerical implementation}

\subsection{Batched finite-element assembly}

For each wavenumber $k_j$ and current source $s$, the 2.5-D finite-element problem is
\begin{equation}
\begin{aligned}
\mathbf A_j(\bm{\sigma})\,
\widehat{\mathbf u}^{\mathrm{sec}}_{j,s}
&=
\mathbf b_{j,s}(\bm{\sigma})
\\
&\equiv
\mathbf b^{\mathrm{ref}}_{j,s}
-
\mathbf A_j(\bm{\sigma})\,
\widehat{\mathbf u}^{\mathrm{pri}}_{j,s}(\bm{\sigma}),
\\
\widehat{\mathbf u}^{\mathrm{pri}}_{j,s}(\bm{\sigma})
&=
\rho_s(\bm{\sigma})\,
\widehat{\mathbf u}^{\mathrm{unit}}_{j,s},
\\
\mathbf A_j(\bm{\sigma})
&=
\sum_{e=1}^{N_c}
\sigma_e
\left(
\mathbf K_e+k_j^2\mathbf M_e
\right)
+
\mathbf B_j(\bm{\sigma}).
\end{aligned}
\label{eq:si_forward_system}
\end{equation}
Here, $\widehat{\mathbf u}^{\mathrm{sec}}_{j,s}$ and $\widehat{\mathbf u}^{\mathrm{pri}}_{j,s}$ are the secondary and primary
potentials, respectively; $\widehat{\mathbf u}^{\mathrm{unit}}_{j,s}$ is the unit-resistivity primary field, $\rho_s(\bm{\sigma})$ is the resistivity evaluated at the source location, and $\mathbf b^{\mathrm{ref}}_{j,s}$ is the fixed reference source term. The effective right-hand side therefore depends on conductivity through both the source resistivity and the primary--secondary correction. The local stiffness and mass matrices are evaluated once from the fixed mesh geometry. In the implementation, element gradients, basis functions and quadrature weights are stored as tensors. The local matrices are obtained for all elements through batched contractions,
\begin{equation}
 K_{eab}=\sum_{q,d}w_{eq}G_{eqad}G_{eqbd},
 \qquad
 M_{eab}=\sum_q w_{eq}N_{qa}N_{qb},
 \label{eq:si_local}
\end{equation}
where $G_{eqad}=\partial_dN_a(\mathbf x_{eq})$. At a model update, the code broadcasts the conductivity and wavenumber dimensions to form
\begin{equation}
 \mathcal A_{jeab}=\sigma_e\left(K_{eab}+k_j^2M_{eab}\right).
 \label{eq:si_wavebatch}
\end{equation}
The Robin boundary contribution is assembled in the same way from precomputed boundary mass matrices and geometry coefficients.

The global sparse pattern is fixed because the mesh topology does not change during inversion. A routing map assigns each local matrix entry to a position in the global CSR matrix. The CSR row pointers and column indices are therefore constructed once; only the nonzero values are updated when $\bm\sigma$ changes. This avoids rebuilding or sorting the sparse graph inside the inversion loop.

\subsection{Sparse factorization and reuse}

AD-TLERT solves the wavenumber-indexed systems with NVIDIA cuDSS. For a given model, each matrix $\mathbf A_j$ is paired with all current sources as multiple right-hand sides,
\begin{equation}
 \mathbf A_j
 \begin{bmatrix}
 \widehat{\mathbf u}_{j,1} & \cdots & \widehat{\mathbf u}_{j,N_s}
 \end{bmatrix}
 =
 \begin{bmatrix}
 \mathbf b_{j,1} & \cdots & \mathbf b_{j,N_s}
 \end{bmatrix}.
 \label{eq:si_mrhs}
\end{equation}
The symbolic plan, sparse indices and allocated work buffers are retained for the fixed mesh and survey. A conductivity update changes the matrix values, so the numerical factors are recomputed. These factors are then reused for all forward, tangent or adjoint right-hand sides associated with the same operator.

Torch and CuPy exchange CUDA-resident arrays through DLPack. No host copy is required when both arrays already reside on the GPU. Geometry-dependent quantities, including element matrices, interpolation operators, source mappings, geometric factors and wavenumber quadrature, are also retained between time steps.

\subsection{Implicit derivatives and batched sensitivities}

The sparse factorization is not differentiated operation by operation. Differentiating equation~\eqref{eq:si_forward_system} gives
\begin{equation}
 \mathbf A_j\,\delta\widehat{\mathbf u}_{j,s}
 =\delta\mathbf b_{j,s}-\delta\mathbf A_j\widehat{\mathbf u}_{j,s}.
 \label{eq:si_jvp}
\end{equation}
For reverse-mode differentiation, the adjoint field satisfies
\begin{equation}
 \mathbf A_j^{\mathsf T}\bm\lambda_{j,s}=\overline{\mathbf u}_{j,s},
 \label{eq:si_adjoint}
\end{equation}
where $\overline{\mathbf u}_{j,s}$ is the cotangent received from the observation and objective functions. The resulting conductivity gradient is
\begin{equation}
 \left.\frac{\partial\Phi}{\partial\sigma_e}\right|_{\mathrm{PDE}}
 =\sum_{j,s}\bm\lambda_{j,s}^{\mathsf T}
 \left(
 \frac{\partial\mathbf b_{j,s}}{\partial\sigma_e}
 -\frac{\partial\mathbf A_j}{\partial\sigma_e}
 \widehat{\mathbf u}_{j,s}
 \right).
 \label{eq:si_vjp}
\end{equation}
Thus, automatic differentiation receives the derivative of the discrete PDE solve without recording the internal sparse ordering and factorization steps.

For the Gauss--Newton experiments, transfer-resistance sensitivities are
evaluated using reciprocity between sources and receivers:
\begin{equation}
\begin{aligned}
 \frac{\partial R_q}{\partial\sigma_e}
 &=-\sum_{j=1}^{N_k}w_j
 \left(\widehat{\mathbf u}_{j,A_q}-\widehat{\mathbf u}_{j,B_q}\right)_e^{\mathsf T}\\
 &\quad\times\left(\mathbf K_e+k_j^2\mathbf M_e\right)
 \left(\widehat{\mathbf u}_{j,M_q}-\widehat{\mathbf u}_{j,N_q}\right)_e.
\end{aligned}
 \label{eq:si_reciprocity}
\end{equation}
A CuPy kernel evaluates this local contraction for blocks of measurements and accumulates the result directly into the requested model parameters. Processing the Jacobian by measurement blocks avoids storing an additional tensor over wavenumber, measurement, cell and local node.

\subsection{Linearised model update}

The Gauss--Newton step is obtained from the augmented least-squares problem
\begin{equation}
 \min_{\delta\mathbf m}
 \left\|
 \begin{bmatrix}
 \mathbf W_d\mathbf J\\
 \sqrt{\alpha_s}\mathbf L_s\\
 \sqrt{\alpha_t}\mathbf L_t
 \end{bmatrix}\delta\mathbf m-
 \begin{bmatrix}
 \mathbf r_d\\
 \mathbf r_s\\
 \mathbf r_t
 \end{bmatrix}
 \right\|_2^2.
 \label{eq:si_augmented}
\end{equation}
The rectangular system is solved with CGLS on the GPU. Matrix-vector products, transpose products and vector updates remain in device memory, and the normal matrix is not formed explicitly.

\begin{table}
\centering
\caption{Numerical settings used for the synthetic windowed inversions and the pyGIMLi benchmark.}
\label{tab:si_settings}
\begin{tabular}{p{0.28\textwidth}p{0.18\textwidth}p{0.42\textwidth}}
\toprule
Setting & Value & Notes \\
\midrule
Precision & float64 & Used in all reported tests \\
Sparse solver & NVIDIA cuDSS & One factorization per model and wavenumber \\
Linear solver & GPU CGLS & Tol. $10^{-12}$; max. 60 iterations \\
Electrodes & 48 & Along the 250-m profile \\
Array & Wenner--alpha & Fixed survey geometry \\
Data per survey & 360 & Same ABMN layout \\
Window length / step & 3 / 1 surveys & 363 windows in total \\
Spatial weight $\alpha_s$ & 50 & First-order smoothness \\
Temporal weight $\alpha_t$ & 10 & Adjacent-time differences \\
Structural weight & 0.05 & Structural prior only \\
Resistivity bounds & $10^{-3}$--$10^4\ \Omega\,\mathrm{m}$ & Applied in resistivity space \\
Max. nonlinear iterations & 15 per window & Early stopping allowed \\
Step tolerance & $10^{-4}$ & Normalized model step \\
Line search & Enabled & Nonlinear updates \\
\bottomrule
\end{tabular}
\end{table}

\section{Objective functions and time-lapse constraints}

\subsection{Data transformation and weighting}

The data misfit is evaluated in natural-log apparent-resistivity space. Let $\epsilon_i$ denote the relative error assigned to datum $i$. The corresponding standard deviation in log-data space is
\begin{equation}
\sigma_{d,i}=\log(1+\epsilon_i),
\qquad
w_{d,i}=\sigma_{d,i}^{-1},
\label{eq:si_data_weight}
\end{equation}
where $w_{d,i}$ is the diagonal entry of the data-weighting matrix $\mathbf{W}_{d,t}$. For survey $t$, the weighted residual is
\begin{equation}
\mathbf{r}_t
=
\mathbf{W}_{d,t}
\left(
\log\bm{\rho}_{a,t}^{\pred}
-
\log\bm{\rho}_{a,t}^{\obs}
\right).
\label{eq:si_weighted_residual}
\end{equation}
A uniform relative error of $\epsilon_i=0.03$ was assigned to all synthetic measurements when constructing the data weights.

\subsection{Data-misfit functions}

The weighted log-$L_2$ objective is
\begin{equation}
 \Phi_d^{L_2}=\sum_t\mathbf r_t^{\mathsf T}\mathbf r_t.
 \label{eq:si_l2}
\end{equation}
The smooth log-$L_1$ objective is
\begin{equation}
 \Phi_d^{L_1}
 =\sum_{t,i}2\left(\sqrt{r_{t,i}^2+\epsilon_d^2}-\epsilon_d\right),
 \qquad \epsilon_d=10^{-3}.
 \label{eq:si_l1}
\end{equation}
The Huber objective is
\begin{equation}
 \Phi_d^{\mathrm H}=\sum_{t,i}h_{\delta_d}(r_{t,i}),
 \qquad \delta_d=1,
 \label{eq:si_huber}
\end{equation}
with
\begin{equation}
 h_{\delta}(r)=
 \begin{cases}
 r^2, & |r|\leq\delta,\\
 2\delta|r|-\delta^2, & |r|>\delta.
 \end{cases}
 \label{eq:si_huber_piece}
\end{equation}
Equations~\eqref{eq:si_l1} and \eqref{eq:si_huber} are solved by IRLS using the residual-dependent weights defined by these objectives.

For the log-data-difference objective, the first survey is fitted in absolute log-data space. Later surveys are fitted relative to that baseline,
\begin{equation}
\begin{aligned}
 \mathbf e_{\Delta,t}
 &=\left(\log\bm{\rho}_{\mathrm a,t}^{\pred}
 -\log\bm{\rho}_{\mathrm a,1}^{\pred}\right)\\
 &\quad-\left(\log\bm{\rho}_{\mathrm a,t}^{\obs}
 -\log\bm{\rho}_{\mathrm a,1}^{\obs}\right),\qquad t>1.
\end{aligned}
 \label{eq:si_diff_residual}
\end{equation}
If $\sigma_{t,i}$ and $\sigma_{1,i}$ are the log-data standard deviations, the difference weight is
\begin{equation}
 w_{\Delta,t,i}=\left(\sigma_{t,i}^2+\sigma_{1,i}^2\right)^{-1/2}.
 \label{eq:si_diff_weight}
\end{equation}
The complete objective is
\begin{equation}
 \Phi_d^{\Delta L_2}=\|\mathbf r_1\|_2^2
 +\sum_{t=2}^{N_t}\|\mathbf W_{\Delta,t}\mathbf e_{\Delta,t}\|_2^2.
 \label{eq:si_diff_phi}
\end{equation}

\subsection{Spatial and temporal regularization}

Taking a triangular inversion mesh as an example, the spatial regularization is formulated as follows. The operator $\mathbf L_s$ contains one row for each interior edge shared by two cells. With the vertical weight set to one in the reported experiments, each row contains equal and opposite coefficients for the two neighbouring cells.

The first-order smoothness objective is
\begin{equation}
 \Phi_s^{\mathrm{FO}}(\mathbf m_t)=\|\mathbf L_s\mathbf m_t\|_2^2.
 \label{eq:si_fo}
\end{equation}
The model-difference formulation applies the same operator to the baseline model and to changes from the baseline,
\begin{equation}
 \Phi_s^{\Delta m}=\|\mathbf L_s\mathbf m_1\|_2^2
 +\sum_{t=2}^{N_t}\|\mathbf L_s(\mathbf m_t-\mathbf m_1)\|_2^2.
 \label{eq:si_modeldiff}
\end{equation}
For the structural prior, $\mathbf L_s$ is replaced by a structure-guided operator. Edges within a hydrostratigraphic unit retain unit weight, whereas edges that cross a unit boundary are assigned a relative weight of 0.05. The spatial Huber objective applies equation~\eqref{eq:si_huber_piece} to $\mathbf L_s\mathbf m_t$ with $\delta_s=1$ and uses IRLS for the local linearization.

The default temporal term is
\begin{equation}
 \Phi_t=\sum_{t=2}^{N_t}\|\mathbf m_t-\mathbf m_{t-1}\|_2^2.
 \label{eq:si_temporal}
\end{equation}
For the direct water-content experiment, the spatial and temporal terms are evaluated for water content rather than resistivity.

\subsection{Sliding windows}

For 365 surveys, window $\ell$ contains
\begin{equation}
 W_\ell=\{\ell,\ell+1,\ell+2\},\qquad \ell=1,\ldots,363.
 \label{eq:si_window}
\end{equation}
The objective within a window is
\begin{equation}
 \Phi_\ell=
 \sum_{t\in W_\ell}\Phi_d^{(t)}
 +\alpha_s\sum_{t\in W_\ell}\Phi_s^{(t)}
 +\alpha_t\sum_{\substack{t,t-1\in W_\ell}}\Phi_t(\mathbf m_t,\mathbf m_{t-1}).
 \label{eq:si_window_phi}
\end{equation}
Each window is inverted independently from the corresponding three initial models. A survey that appears in more than one window receives multiple estimates. These estimates are averaged in natural-log resistivity space,
\begin{equation}
 \overline{\mathbf m}_t
 =\frac{1}{n_t}\sum_{\ell:t\in W_\ell}\mathbf m_t^{(\ell)},
 \label{eq:si_window_average}
\end{equation}
which is equivalent to a geometric mean in resistivity space. For a petrophysical parameterization, the final parameter field is obtained from the aggregated log-resistivity model through the same differentiable mapping.

\section{Synthetic hydrologic and petrophysical models}

\subsection{ParFlow--CLM configuration}

The synthetic hillslope is 250 m long, 1 m wide and approximately 60 m deep. The ParFlow grid contains 50 cells in the profile direction, five cells across strike and 25 non-uniform vertical layers. The horizontal spacings are 5 m and 0.2 m. The nominal vertical spacing is 10 m and is multiplied by the layer-specific scaling factors in the public configuration file. The water-year 2023 forcing is applied at hourly resolution.

The subsurface contains regolith, fractured bedrock and fresh bedrock. The first two units use truncated lognormal Saturated hydraulic conductivity fields generated by the turning-bands method. Their geometric means differ by two orders of magnitude, while the fresh-bedrock value is constant. The random fields use a standard deviation of 0.6 and correlation lengths of 50, 1 and 5 m in the $x$, $y$ and $z$ directions, respectively. The configured unit properties are listed in Table~\ref{tab:si_hydrology}.

\begin{center}
\begin{minipage}{0.96\textwidth}
\centering
\footnotesize

\captionof{table}{Hydrologic parameters used in the ParFlow--CLM model. The Saturated hydraulic conductivity column gives the configured geometric mean for the random fields and the constant value for fresh bedrock.}
\label{tab:si_hydrology}

\begin{tabular*}{\linewidth}{@{\extracolsep{\fill}}lcccccc@{}}
\toprule
Unit & Porosity & Saturated hydraulic  & Specific storage
& $\alpha$ & $n$ & Residual saturation \\
& & conductivity (m h$^{-1}$) & (m$^{-1}$) & & & \\
\midrule
Regolith
& 0.35
& 3.6
& $5\times10^{-6}$
& 3.5
& 2.0
& 0.10 \\

Fractured bedrock
& 0.27
& $3.6\times10^{-2}$
& $1\times10^{-6}$
& 3.5
& 2.0
& 0.10 \\

Fresh bedrock
& 0.09
& $3.6\times10^{-4}$
& $5\times10^{-7}$
& 0.1
& 1.5
& 0.10 \\
\bottomrule
\end{tabular*}

\vspace{1.8em}

\captionof{table}{Petrophysical parameters used to construct the synthetic resistivity models and in the direct water-content inversion.}
\label{tab:si_petrophysics}

\begin{tabular*}{\linewidth}{@{\extracolsep{\fill}}lcccccc@{}}
\toprule
Unit & $\phi$ & $n$ & $\rho_{\mathrm{sat}}$
& $\rho_{\mathrm{sat,s}}$ & $\sigma_p$ & $\sigma_s$ \\
& & & ($\Omega\,\mathrm{m}$)
& ($\Omega\,\mathrm{m}$)
& (S m$^{-1}$)
& (S m$^{-1}$) \\
\midrule
Regolith
& 0.35
& 2.2
& 170
& 510
& $3.9216\times10^{-3}$
& $1.9608\times10^{-3}$ \\

Fractured bedrock
& 0.27
& 1.8
& 1100
& --
& $9.0909\times10^{-4}$
& 0 \\

Fresh bedrock
& 0.09
& 2.5
& 2400
& --
& $4.1667\times10^{-4}$
& 0 \\
\bottomrule
\end{tabular*}

\end{minipage}
\end{center}

\subsection{Petrophysical parameters}

The synthetic resistivity sequence is generated with the unit-dependent relationship
\begin{equation}
 \sigma(S)=\sigma_pS^n+\sigma_sS^{n-1},
 \qquad \rho(S)=\sigma(S)^{-1}.
 \label{eq:si_petrophysics}
\end{equation}
The preset stores the saturated resistivity $\rho_{\mathrm{sat}}$ and, where surface conduction is included, $\rho_{\mathrm{sat,s}}$. The coefficients used in equation~\eqref{eq:si_petrophysics} are
\begin{equation}
 \sigma_s=\rho_{\mathrm{sat,s}}^{-1},
 \qquad
 \sigma_p=\rho_{\mathrm{sat}}^{-1}-\sigma_s.
 \label{eq:si_sigma_coeffs}
\end{equation}
When $\rho_{\mathrm{sat,s}}$ is absent, $\sigma_s=0$ and $\sigma_p=\rho_{\mathrm{sat}}^{-1}$. Porosity is read from the ParFlow model. The parameters mapped to the inversion mesh are given in Table~\ref{tab:si_petrophysics}. A global saturation floor of $10^{-4}$ is used in the bounded parameterization.
The same parameter fields are used to generate the synthetic resistivity models and to define the direct petrophysical inversion. The comparison with post-inversion conversion therefore isolates the effect of the inversion parameterization and the space in which regularization is applied.

\section{Cross-code comparisons}

The main text already benchmarks AD-TLERT against pyGIMLi. To provide additional tests of numerical consistency and computational efficiency, we further compare AD-TLERT with ResIPy \citep{Blanchy2020} and ADERT \citep{Rincon2026}. ResIPy generates a substantially denser inversion mesh under its default settings. A runtime comparison would therefore reflect differences in both mesh discretisation and solver implementation, so the comparison with ResIPy is restricted to the accuracy of the recovered resistivity structures and time-lapse changes. ADERT provides a more direct efficiency benchmark because both ADERT and AD-TLERT use automatic differentiation. ADERT implements matrix assembly, the PDE solution and data projection within the PyTorch computational graph, and the GPU reference implementation tested here uses an iterative PCG solver \citep{Rincon2026}. In contrast, AD-TLERT differentiates the sparse PDE solve implicitly and uses cuDSS sparse direct factorization, multiple right-hand sides and factorization reuse across forward and gradient calculations. This design is intended for repeated 2.5-D time-lapse calculations and reduces both repeated-solve overhead and GPU memory use. The ADERT benchmark therefore compares numerical consistency, forward and gradient runtimes, and peak GPU memory on a common reference model.

\subsection{ResIPy comparison}

Both implementations use the same synthetic observations and three-survey sliding-window layout. Their inversion meshes are generated independently, so the recovered models are interpolated in log-resistivity space to the centers of the common truth cells before the errors are evaluated. The comparison is therefore used to assess the recovery of the main spatial structures and time-lapse changes. Differences between the results may reflect both the inversion formulation and the independent mesh discretizations.

\begin{figure}
\centering
\includegraphics[width=\textwidth]{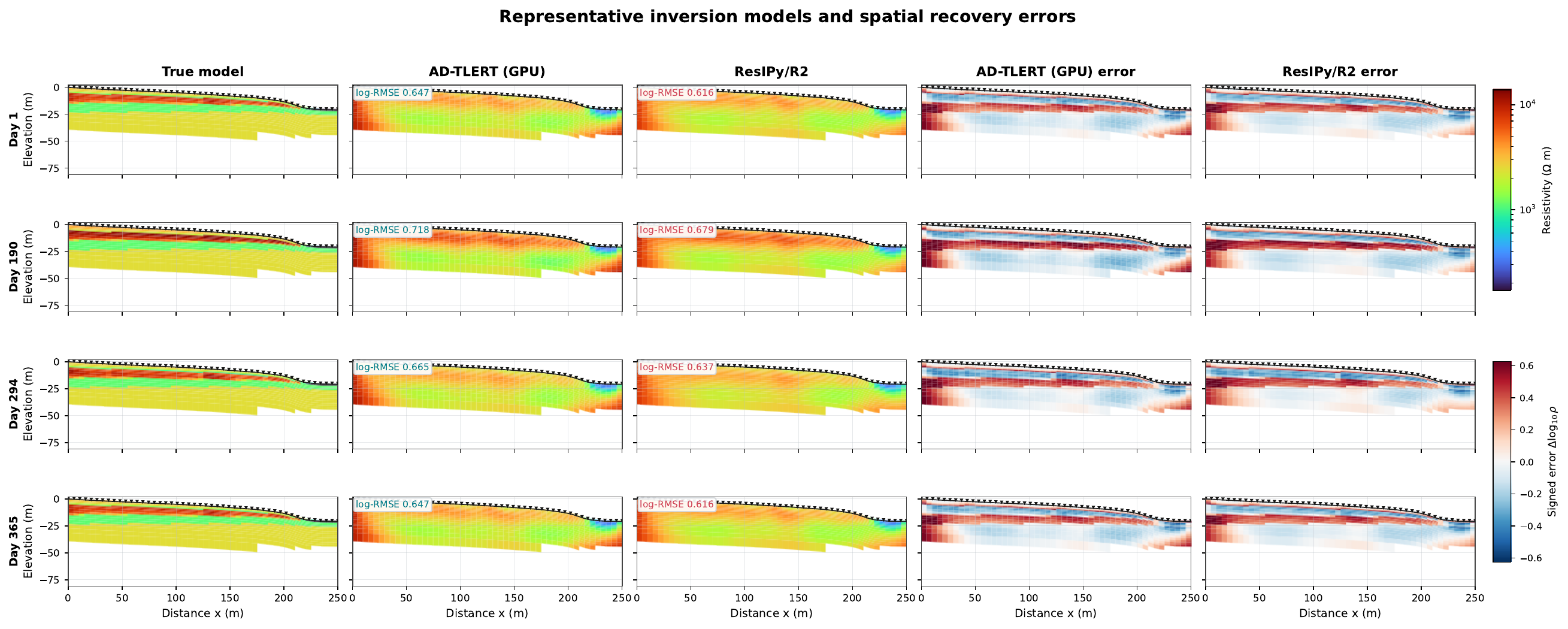}
\caption{Representative resistivity models and spatial recovery errors for Days 1, 190, 294 and 365. From left to right, the columns show the true resistivity model, the AD-TLERT result, the ResIPy result, the signed AD-TLERT error and the signed ResIPy error. The recovered models were interpolated in log-resistivity space to the same truth-cell centers within their shared support.}
\label{fig:si_resipy_models}
\end{figure}

Both implementations recover the main resistivity structure of the synthetic hillslope model, including the resistive shallow layer and the lower-resistivity material at depth and near the downslope end (Fig.~\ref{fig:si_resipy_models}). The spatial distributions of the errors are also similar, with the largest discrepancies concentrated near the shallow resistivity contrasts and along the boundaries of the resolved model domain. ResIPy gives slightly lower absolute-model $\ln\rho$ RMSE values at the four representative times. Its values range from 0.616 to 0.679, compared with 0.647 to 0.718 for AD-TLERT. This modest difference may partly reflect the denser inversion mesh generated by the default ResIPy configuration and the differences in regularization
and discretization between the two implementations.

\begin{figure}
\centering
\includegraphics[width=\textwidth]{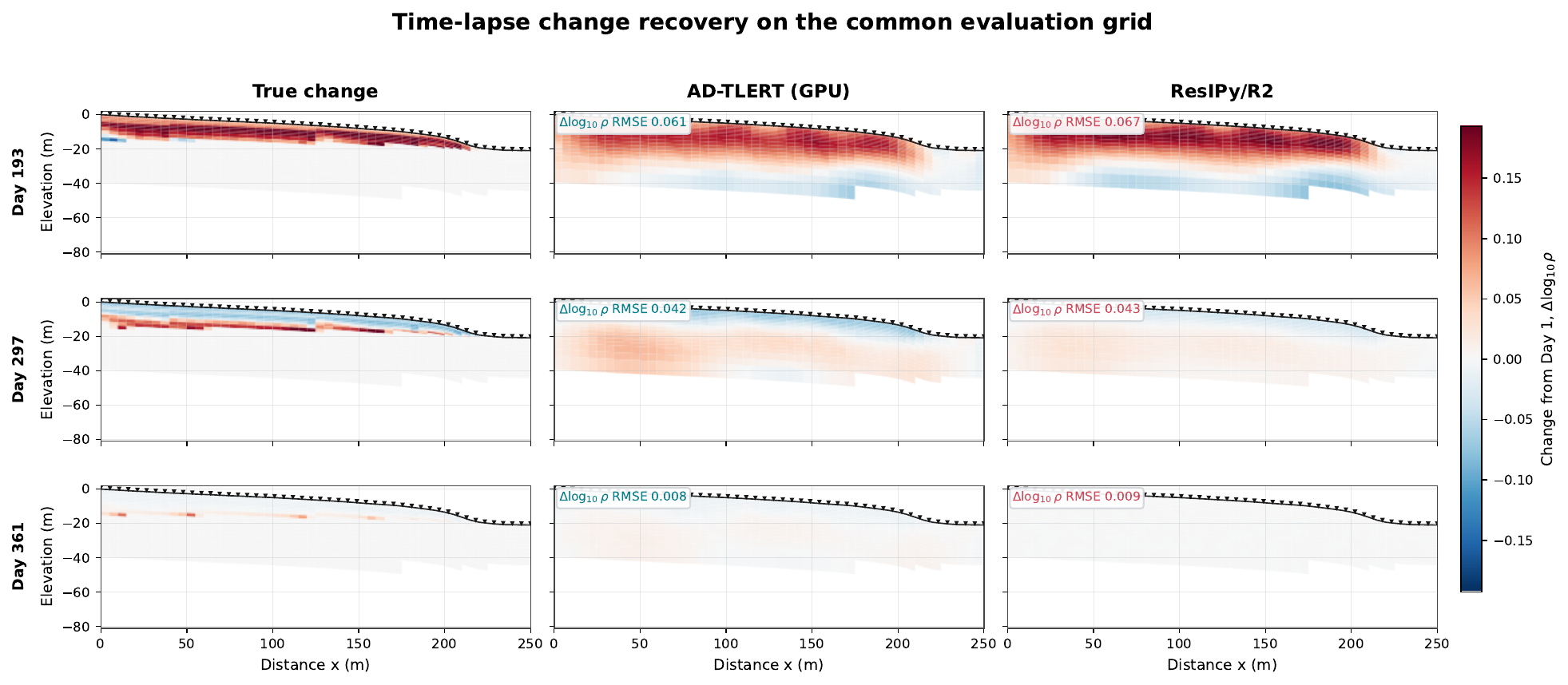}
\caption{Recovery of time-lapse resistivity changes relative to Day 1. The rows show Days 193, 297 and 361, which are the available surveys closest to Days 190, 294 and 365. From left to right, the columns show the true change, the AD-TLERT result and the ResIPy result. True and recovered changes are evaluated on the same grid and displayed with a common colour scale. }
\label{fig:si_resipy_changes}
\end{figure}

The agreement is closer when the comparison is made in terms of time-lapse changes (Fig.~\ref{fig:si_resipy_changes}). For Days 193, 297 and 361, the $\Delta\ln\rho$ RMSE values are 0.061, 0.042 and 0.008 for AD-TLERT, and 0.067, 0.043 and 0.009 for ResIPy, respectively. Both methods recover the stronger shallow changes during the main wetting period and the subsequent reduction in change amplitude. AD-TLERT gives slightly lower change errors at all three selected times, although the differences between the two methods are small. The comparison therefore shows that AD-TLERT produces time-lapse changes consistent with an established independent inversion code. It is not intended as a strict ranking of the two methods because their meshes and numerical
formulations are not identical.

\subsection{ADERT reference-model benchmark}

AD-TLERT was also compared with the published ADERT implementation \citep{Rincon2026} using the official ADERT 2.5-D reference model. The two calculations used the same 6,240 active parameters, 8,816 forward-model cells, 9,087 nodes and 195 dipole--dipole measurements. Each calculation was warmed up once and then repeated three times; the median elapsed time is reported.

\begin{table}
\centering
\caption{GPU performance on the ADERT 2.5-D reference model. Peak memory includes the CUDA context and solver state for the complete process.}
\label{tab:si_adert_benchmark}

\begin{tabular*}{\textwidth}{@{\extracolsep{\fill}}lrrrr@{}}
\toprule
Implementation
& Forward (s)
& Gradient (s)
& Total (s)
& Peak memory (MiB) \\
\midrule
AD-TLERT cuDSS
& 0.0783
& 0.1874
& 0.2656
& 485 \\
ADERT PCG
& 1.4328
& 1.3066
& 2.7350
& 961 \\
\bottomrule
\end{tabular*}

\end{table}

The two automatic-differentiation implementations produce closely consistent numerical results. The relative $L_2$ difference between the predicted data is 2.15 per cent, while the correlation coefficient between the logarithmic predictions is 0.9970. The gradients also agree closely within the active inversion region, with a correlation coefficient of 0.9965 and a cosine similarity of 0.9982. These results indicate that the two implementations propagate comparable sensitivity information despite differences in their linear solvers and wavenumber discretizations.

The main difference is computational performance. AD-TLERT completes the forward calculation in 0.0783~s and the gradient calculation in 0.1874~s, compared with 1.4328 and 1.3066~s, respectively, for the tested ADERT GPU implementation. The combined forward-and-gradient calculation is therefore approximately 10.3 times faster with AD-TLERT. The full-process peak GPU memory is also reduced from 961~MiB to 485~MiB. This improvement is associated with the use of cuDSS sparse direct factorization, multiple right-hand-side solves, and reuse of the matrix structure and numerical factors, while the sparse PDE derivative is supplied through implicit differentiation.

\section{Field-data processing and auxiliary observations}

\subsection{Temperature model and correction}

The field example follows the temperature treatment used in the original field study of Uhlemann et al. (2024). Seasonal ground temperature at depth $z$ and day of year $t$ is represented by
\begin{equation}
 T_{\mathrm{model}}(z,t)=T_{\mathrm{mean}}
 +\frac{\Delta T}{2}\exp\left(-\frac{z}{d}\right)
 \sin\left(\frac{2\pi t}{365}+\Phi-\frac{z}{d}\right),
 \label{eq:si_temperature}
\end{equation}
where $T_{\mathrm{mean}}$ is the mean annual ground-surface temperature, $\Delta T$ is the annual temperature range, $d$ is the characteristic decay depth, $z$ is positive downward from the ground surface and $\Phi$ is the phase offset. Equation~\eqref{eq:si_temperature} is fitted at the temperature-sensor locations and interpolated over the ERT domain.

The ratio correction to the reference temperature is
\begin{equation}
 \rho_{\mathrm{corr}}=\rho\,C_T,
 \qquad
 C_T=1+\frac{c_t}{100}
 \left(T_{\mathrm{ref}}-T_{\mathrm{model}}\right),
 \label{eq:si_temperature_correction}
\end{equation}
with reference temperature $T_{\mathrm{ref}}=6.5\ ^\circ\mathrm C$ and $c_t=2.0$ per cent $^\circ\mathrm C^{-1}$. In the differentiable field parameterization, the resistivity at survey temperature is written as
\begin{equation}
 \log\rho_t=\log\rho_0
 -n\log\left(\frac{\theta_t}{\theta_0}\right)
 -\log C_T(t),
 \label{eq:si_relative_archie}
\end{equation}
where $\rho_0$ and $\theta_0$ are the baseline resistivity and water-content models at the reference state.

\subsection{TMC observations and soft constraint}

Six monitoring stations are located 18, 62, 124, 148, 195 and 231 m along the transect. Soil moisture is measured at depths of 0.1, 0.3, 0.6 and 0.8 m every 15 min and averaged to hourly values. Co-located temperature strings, except at 124 m, extend to approximately 1 m depth. Temperature observations define $C_T$ in equation~\eqref{eq:si_temperature_correction}; the TMC water-content observations enter the inversion as a separate soft constraint.

Let $\mathbf P_c$ map the gridded water-content model to the sensor locations and depths, and let $\bm\theta_t^{\mathrm{TMC}}$ contain the processed sensor observations. The constraint is
\begin{equation}
 \Phi_c=\sum_t
 \left\|\mathbf W_{c,t}
 \left(\mathbf P_c\bm\theta_t-\bm\theta_t^{\mathrm{TMC}}\right)
 \right\|_2^2,
 \label{eq:si_sensor_constraint}
\end{equation}
where $\mathbf W_{c,t}$ contains the sensor weights. Its contribution to the full objective is controlled by $\alpha_c$. The sensor term is assembled in the physical water-content domain and is differentiated through the same parameter mapping as the ERT data term.

\nocite{Uhlemann2024}
\bibliographystyle{ad_tlert_references}
\bibliography{adert_references}